\documentclass{article}

\usepackage{microtype}
\usepackage{graphicx}
\usepackage{subfigure}
\usepackage{booktabs} % for professional tables

\usepackage{caption}

\usepackage{hyperref}

\usepackage{amsmath}
\usepackage{amssymb}
\usepackage{mathtools}
\usepackage{amsthm}

\usepackage{amsmath,amsfonts,bm}

\def\eqref#1{(\ref{#1})}
\def\Eqref#1{(\ref{#1})}
\def\1{\bm{1}}

\DeclareMathAlphabet{\mathsfit}{\encodingdefault}{\sfdefault}{m}{sl}
\SetMathAlphabet{\mathsfit}{bold}{\encodingdefault}{\sfdefault}{bx}{n}

\def\gB{{\mathcal{B}}}
\def\gC{{\mathcal{C}}}
\def\gD{{\mathcal{D}}}

\def\gF{{\mathcal{F}}}
\def\gG{{\mathcal{G}}}

\def\gM{{\mathcal{M}}}

\def\gO{{\mathcal{O}}}

\def\gT{{\mathcal{T}}}

\def\gW{{\mathcal{W}}}
\def\gX{{\mathcal{X}}}
\def\gY{{\mathcal{Y}}}

\newcommand{\E}{\mathbb{E}}

\newcommand{\R}{\mathbb{R}}

\DeclareMathOperator*{\argmin}{arg\,min}

\usepackage[accepted]{icml2024}

\usepackage[capitalize,noabbrev]{cleveref}

\theoremstyle{plain}
\newtheorem{theorem}{Theorem}[section]
\newtheorem{proposition}[theorem]{Proposition}

\theoremstyle{definition}

\theoremstyle{remark}

\usepackage[textsize=tiny]{todonotes}

\usepackage{accents}
\newcommand{\ubar}[1]{\underaccent{\bar}{#1}}

\icmltitlerunning{Importance Weighted Group Accuracy Estimation for Improved Calibration and Model Selection}

\begin{document}

\twocolumn[
\icmltitle{IW-GAE: Importance Weighted Group Accuracy Estimation for Improved Calibration and Model Selection in Unsupervised Domain Adaptation}




\begin{icmlauthorlist}
\icmlauthor{Taejong Joo}{nu}
\icmlauthor{Diego Klabjan}{nu}
\end{icmlauthorlist}

\icmlaffiliation{nu}{Department of Industrial Engineering \& Management Sciences, Northwestern University, Evanston, IL, USA}

\icmlcorrespondingauthor{Taejong Joo}{taejong.joo@northwestern.edu}
\icmlcorrespondingauthor{Diego Klabjan}{d-klabjan@northwestern.edu}

\icmlkeywords{Machine Learning, ICML}

\vskip 0.3in
]



\printAffiliationsAndNotice{}

\begin{abstract}
    Distribution shifts pose significant challenges for model calibration and model selection tasks in the unsupervised domain adaptation problem---a scenario where the goal is to perform well in a distribution shifted domain without labels. In this work, we tackle difficulties coming from distribution shifts by developing a novel importance weighted group accuracy estimator. Specifically, we present a new perspective of addressing the model calibration and model selection tasks by estimating the group accuracy. Then, we formulate an optimization problem for finding an importance weight that leads to an accurate group accuracy estimation with theoretical analyses. Our extensive experiments show that our approach improves state-of-the-art performances by 22\% in the model calibration task and 14\% in the model selection task. 
\end{abstract}

\section{Introduction}
In this work, we consider a classification problem in unsupervised domain adaptation (UDA). UDA aims to transfer knowledge from a source domain with ample labeled data to enhance the performance in a target domain where labeled data is unavailable. In UDA, the source and target domains have different data generating distributions, so the core challenge is to transfer knowledge contained in the labeled dataset in the source domain to the target domain under the distribution shifts. Over the decades, significant improvements in the transferability of accuracy from source to target domains have been made, resulting in areas like domain alignment \citep{ben2010theory,zhang2019bridging} and self-training \citep{chen2020self,cai2021theory}.

\begin{figure*}
    \vskip -0.10in
     \centering{
     \hfill
     \begin{subfigure}[]
         {\includegraphics[width=0.39\textwidth]{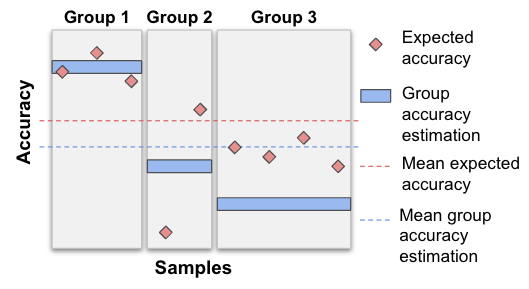}
         \label{fig:figure1b}}
     \end{subfigure}
     \hfill
     \begin{subfigure}[]
         {\includegraphics[width=0.29\textwidth]{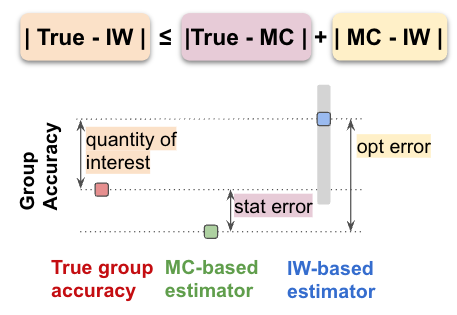}
         \label{fig:figure1a}}
     \end{subfigure}
     \hfill}
    \vskip -0.15in
    \caption{
    Figure \ref{fig:figure1b} illustrates ideal and failure cases of IW-GAE with nine data points (red diamonds) from three groups (gray boxes). Group 1 is desirable for model calibration where the group accuracy estimation (a blue rectangle) well represents the individual expected accuracies of samples in the group. Conversely, group accuracy estimation could inaccurately represent the individual accuracies in the group due to a high variance of accuracies within the group (group 2) and a high bias of the estimator (group 3). 
    For model selection, we aim to match the mean group accuracy estimation (the blue dotted line as an average of blue rectangles) to the mean expected accuracy (the red dotted line as an average of red diamonds), which can be induced by accurate group accuracy estimations for each group.
    Figure \ref{fig:figure1a} explains the idea of encouraging two estimators close to each other. The shaded area for the IW-based estimator is possible group accuracy estimations from different IWs. IW-GAE finds the IW minimizing the opt error for reducing the group accuracy estimation error.}
    \label{fig:figure1}
    \vskip -0.15 in
\end{figure*}

However, model calibration, which is about matching prediction confidence on a sample to its expected accuracy \citep{dawid1982well,guo2017calibration}, remains challenging in UDA due to the distribution shifts. Specifically, it is widely known that state-of-the-art calibrated classifiers in the independent and identically distributed (i.i.d.) settings \citep{guo2017calibration,gal2016dropout,lakshminarayanan2017simple} begin to generate over-confident predictions in the face of distributional shifts \citep{ovadia2019can}. Further, \citet{wang2020transferable} show the discernible compromise in calibration performance as an offset against the enhancement of the accuracy in the target domain.

Moreover, the model selection task in UDA remains challenging due to the scarcity of labeled target domain data that are required to evaluate model performance. In the i.i.d. settings, a standard approach for model selection is a cross-validation method---constructing a hold-out dataset for selecting the model that yields the best performance on the hold-out dataset. While cross-validation provides favorable statistical guarantees  \citep{stone1977asymptotics,kohavi1995study}, such guarantees falter in the presence of the distribution shifts due to the violation of the i.i.d. assumption. In practice, it has also been observed that performances of machine learning models measured in one domain have significant discrepancies to their performances in another distribution shifted domain \citep{hendrycks2019benchmarking,ovadia2019can,recht2019imagenet}. Therefore, applying model selection techniques in the i.i.d. settings to the labeled source domain is suboptimal in the target domain.

In this work, we simultaneously address these critical aspects in UDA from a new perspective of predicting a group accuracy.
Specifically, we partition predictions into a set of groups and then estimate the group accuracy---the average accuracy of predictions in a group. When the group accuracy estimate accurately represents the expected accuracy of a model for samples in the group (e.g., group 1 in Figure~\ref{fig:figure1b}), using the group accuracy estimate as prediction confidence induces a well-calibrated classifier. When the average of the group accuracy estimates matches the mean expected accuracy (e.g., two dotted lines in Figure \ref{fig:figure1b} are close to each other), it becomes a good model selection criterion.

To this end, we propose \textbf{importance weighted group accuracy estimation (IW-GAE)} that aims to find importance weights (IWs) that induce an accurate group accuracy estimator under the distribution shifts. Specifically, we define two estimators for the group accuracy in the source domain (MC-based and IW-based estimators in Figure \ref{fig:figure1a}), where only one of them depends on the IW. Then, we formulate a novel optimization problem for finding the IW that makes the two estimators close to each other (reducing opt error in Figure \ref{fig:figure1a}). Through theoretical analyses and several experiments, we show that the optimization process results in an accurate group accuracy estimator for the target domain (small quantity of interest in Figure \ref{fig:figure1a}), improving model calibration and model selection performances.

Our contributions can be summarized as follows:
1) We show when and why considering group accuracy, instead of the accuracy for individual samples, is statistically favorable, which can simultaneously benefit model calibration and model selection with attractive properties;
2) We propose a novel optimization problem for IW estimation that directly reduces group accuracy estimation error in UDA with theoretical analyses; 
3) On average, IW-GAE improves state-of-the-art performances by 22\% in the model calibration task and 14\% in the model selection task.

\textbf{Notation and problem setup }
Let $\gX \subseteq \R^r$ and $\gY = [K] := \{1, 2, \cdots, K \}$ be input and label spaces. Let $\hat{Y}: \gX \rightarrow [K]$ be the prediction function of a model and $Y(x)$ is a $K$-dimensional categorical random variable related to a label at $X = x$. When there is no ambiguity, we represent $Y(x)$ and $\hat{Y}(x)$ as $Y$ and $\hat{Y}$ for brevity. We are given a labeled source dataset $\gD_S = \{(x^{(S)}_i, y^{(S)}_i)\}_{i=1}^{N^{(S)}}$ sampled from $p_{S_{XY}}$ and an unlabeled target dataset $\gD_T = \{x^{(T)}_i \}_{i \in [N^{(T)}]}$ sampled from $p_{T_X}$ where $p_{S_{XY}}$ is a joint data generating distribution of the source domain and $p_{T_X}$ is a marginal distribution of the target domain. We also denote $\E_p[\cdot]$ as the population expectation and $\hat{\E}_{p}[\cdot]$ as its empirical counterpart. For $p_{S_{XY}}$ and $p_{T_{XY}}$, we consider a covariate shift without a concept shift; i.e., $p_{S_X}(x) \neq p_{T_X}(x)$ but $p_{S_{Y|X}}(y|x) = p_{T_{Y|X}}(y|x)$ for $x \in \gX$. For the rest of the paper, we use the same notation $p_S$ for marginal distribution $p_{S_X}$ and joint distribution $p_{S_{XY}}$ when there is no ambiguity. However, we use the explicit notation for $p_{S_{Y|X}}$ and $p_{T_{Y|X}}$ to avoid confusion.

\section{Group Accuracy Estimation for Model Calibration and Selection}
We address model calibration and model selection tasks in UDA by estimating the group accuracy. Specifically, we construct $M$ groups $\{ \gG_n \}_{n \in [M]}$ with some grouping function $I^{(g)}: \mathcal{X} \rightarrow [M]$. Then, for each group $\gG_n$, we estimate the average accuracy of target domain samples in $\gG_n$ defined as $\alpha_T(\gG_n) := \E_{p_T}[ \textbf{1}(Y(X) = \hat{Y}(X)) | \gG_n ]$. In the following, we first give a motivation for estimating the group accuracy, instead of an expected accuracy for each sample. Then, we explain how to construct groups and use the group accuracy estimates for simultaneously solving model calibration and selection tasks.

\subsection{Motivation for Estimating the Group Accuracy}
Suppose we are given samples $D := \{(x_i, y_i)  \in \gG_n \}_{i \in [N_n]}$ and a classifier $f$. Let $\beta(x_i) := \E_{Y|X = x_i} [\textbf{1}(Y(x_i) = f(x_i))]$ be an expected accuracy of $f$ at $x_i$, which is our goal to estimate. Then, due to realization of a single label at each point, the observed accuracy $\hat{\beta}(x_i) := \textbf{1}(y_i = f(x_i))$ is a random sample from the Bernoulli distribution with parameter $\beta(x_i)$ that has a variance of $\sigma_{x_i}^2 = \beta(x_i) (1 - \beta(x_i))$. Note that this holds when $x_i \neq x_j$ for $i \neq j$, which is the case for most machine learning scenarios. 

Under this setting, we show the sufficient condition that the maximum likelihood estimator (MLE) of the group accuracy outperforms the MLE of the individual accuracy.

\begin{proposition} \label{prop:suff_gae}
    Let $\hat{\beta}^{(id)}$ and $\hat{\beta}^{(gr)}$ be MLEs of individual and group accuracies. Then, $\hat{\beta}^{(gr)}$ has a lower expected mean-squared error than $\hat{\beta}^{(id)}$ if 
    \begin{equation} \label{eq:stat_favorable}
        \tfrac{1}{4} ( \max_{x^\prime \in \gG_n} \beta(x^\prime) - \min_{x^\prime \in \gG_n} \beta(x^\prime)  )^2 \leq \tfrac{N_n - 1}{N_n} \bar{\sigma}^2   
    \end{equation}
    where $\bar{\sigma}^2 = \tfrac{1}{N_n} \sum_{i \in [N_n]} \sigma^2_{x_i}$ with $\sigma^2_{x_i} = \beta(x_i) (1-\beta(x_i))$.
\end{proposition}

The proof is based on bias-variance decomposition and the Popoviciu's inequality \citep{popoviciu1965certaines}, which is given in Appendix \ref{appx:proof_suff}. In Proposition~\ref{prop:suff_gae}, \eqref{eq:stat_favorable} is the condition under which the group accuracy estimator achieves a lower mean-squared error than the individual accuracy estimator. 
Crucially, we can reduce $\max_{x \in \gG_n} \beta(x) - \min_{x \in \gG_n} \beta(x)$ through a careful group construction that we discuss in Section \ref{sec:gr_assign}. Further, a sufficient condition for \eqref{eq:stat_favorable} tends to be loose (cf. Appendix \ref{appx:suff_conf_stat_favorable}). Therefore, under the loose condition on the group construction, \textit{the group accuracy estimator would be statistically more favorable}.

\subsection{Group Assignment} \label{sec:gr_assign}
As seen in Figure~\ref{fig:figure1b}, it is important to construct groups such that the expected accuracy of samples in the same group has a low variance. Therefore, we group examples by the maximum value of the softmax output as in \citet{guo2017calibration} based on an observation that the maximum value of the softmax output is highly correlated with accuracy in UDA \citep{wang2020transferable}. In addition, we note the result that an overall scale of the maximum value of the softmax output significantly varies from one domain to another \citep{yu2022robust}. Thus, we adjust the sharpness of the softmax output in the target domain by introducing a \textit{learnable} temperature parameter $t \in \gT $ where $\gT$ is a bounded interval in $\R_+$.

Equipped with these ideas, we first gather a set of prediction confidences under a temperature $t$, denoted as $\gC^{(t)} := \{ m(x_S; 1) | (x_S, y_S) \in \gD_S \} \cup \{ m(x_T; t) | x_T \in \gD_T \}$ where $m(x; t)$ is the maximum value of the softmax output under $t$ ($t=1$ recovers the standard softmax). Then, we construct the $n$-th confidence group under $t$ for $n \in [M]$ by
\begin{multline}
    \gG_n^{(t)} := \{x \in \gD_S | q(\tfrac{n-1}{M}; \gC^{(t)}) \leq m(x; 1) < q(\tfrac{n}{M}; \gC^{(t)}) \} \\ \cup \{x \in \gD_T | q(\tfrac{n-1}{M}; \gC^{(t)}) \leq m(x; t) < q(\tfrac{n}{M}; \gC^{(t)}) \} 
\end{multline}
where $q(\tfrac{n}{M}; \gC^{(t)})$ is the $\tfrac{n}{M}$-th quantile of $\gC^{(t)}$. For the rest of the paper, we let $I^{(g)}(x_S) = \min \{ k \in [M] : m(x_S;1) < q(\tfrac{k}{M}; \gC^{(t)})\} $ if $x_S \sim p_S$ and $I^{(g)}(x_T) = \min \{ k \in [M]: m(x_T;t) < q(\tfrac{k}{M}; \gC^{(t)})\} $ if $x_T \sim p_T$.

\subsection{Model Selection and Model Calibration}
Before explaining 
how to obtain an accurate group accuracy estimator $\hat{\alpha}_T(\gG_n^{(t)})$ in Section \ref{sec:new_method}, we first show how to use $\hat{\alpha}_T(\gG_n^{(t)})$ to simultaneously solve model calibration and model selection tasks with attractive properties.

\textbf{Model calibration } 
We use $\hat{\alpha}_T(\gG^{(t)}_{I^{(g)}(x)})$ as an estimate of prediction confidence on $x \sim p_T$. Then, we can address the model calibration task in UDA with a bounded calibration error. Specifically, an expected squared calibration error can be decomposed as the sum of the variance of the accuracies for samples in the same group and a squared group accuracy estimation error (cf. Proposition  \ref{prop:brier_score_decomposition}); that is, 
$\E_{p_T}[(P(Y = \hat{Y}) - \hat{\alpha}_T(\gG^{(t)}_{I^{(g)}(X)}))^2] = \sum_{n\in[M]} \E_{p_T}[\textbf{1}(X \in \gG_n^{(t)})] \cdot (Var(P(Y = \hat{Y})|\gG_n^{(t)}) + (\alpha_T(\gG^{(t)}_{n}) - \hat{\alpha}_T(\gG^{(t)}_{n}))^2 )$. 
Thus, combined with the guarantees about the group accuracy estimation error (cf. Proposition \ref{proposition:closeness_opt} and \eqref{eq:target_acc_estim_bound}), our approach can enjoy the bounded calibration error unlike previous approaches using $m(x;t)$ as the prediction confidence estimate \citep{park2020calibrated,wang2020transferable}.

\textbf{Model selection }
We use the average group accuracy estimate $\E_{p_T}[\hat{\alpha}_T(\gG^{(t)}_{I^{(g)}(X)})]$ computed with a hold-out target domain dataset as the model selection criteria. Again, this criteria estimates the average accuracy of the model with a bounded error due to the Cauchy-Schwarz inequality; that is, $|\E_{p_T}[P(Y=\hat{Y})] - \E_{p_T}[\hat{\alpha}_T(\gG^{(t)}_{I^{(g)}(X)})]| \leq ( \E_{p_T}[(P(Y = \hat{Y}) - \hat{\alpha}_T(\gG^{(t)}_{I^{(g)}(X)}))^2] )^{1/2}$. In addition, compared to the approaches \citep{sugiyama2007covariate,you2019towards} aiming to estimate only the mean accuracy in $p_T$, our approach will have an additional regularization effect from encouraging accurate group accuracy estimation for each group.

\begin{figure*}
    \vskip -0.1in
     \centering
     \hfill
     \begin{subfigure}[]
         {\includegraphics[width=0.27\textwidth]{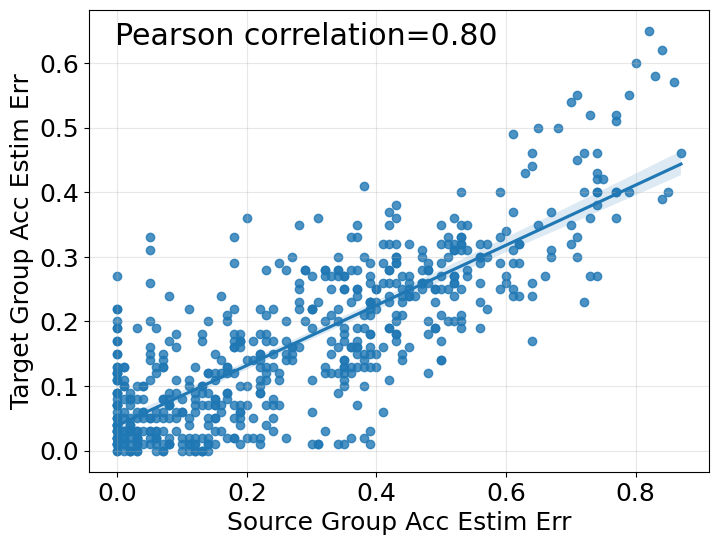}
         \label{fig:aa}}
     \end{subfigure}
     \hfill
     \begin{subfigure}[]
         {\includegraphics[width=0.27\textwidth]{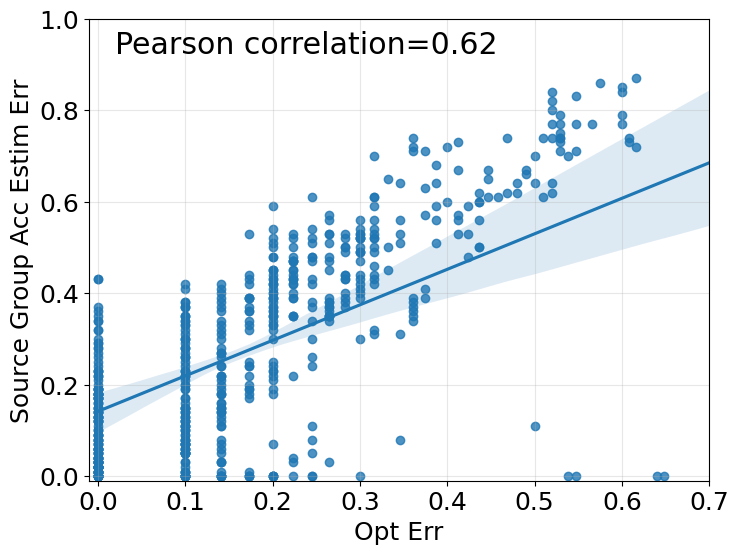}
         \label{fig:bb}}
     \end{subfigure}
     \hfill
     \begin{subfigure}[]
         {\includegraphics[width=0.27\textwidth]{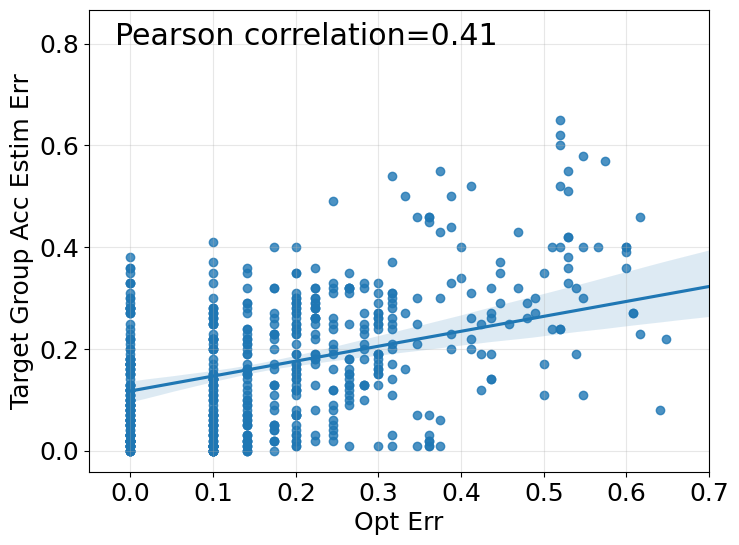}
         \label{fig:cc}}
     \end{subfigure}
     \vskip -0.15in
    \caption{
    Illustration of correlations between the optimization error and the source and target group accuracy estimation errors. Each point corresponds to a different IW estimator and the values are measured on the OfficeHome dataset (720 IW estimators in total). See Appendix \ref{appx:terms_analy} for more detailed discussions and analyses. }
    \label{fig:figure_new_anal}
    \vskip -0.15in
\end{figure*}

\section{Accurate Group Accuracy Estimation via IW-GAE} \label{sec:new_method}
In this section, we propose IW-GAE that aims to accurately estimate ${\alpha}_T(\gG_n^{(t)})$ by using a novel idea tailored for UDA where $Y(x)$ is available for $x \sim p_S$ but not for $x \sim p_T$. A core idea behind IW-GAE is to use importance weighting, which is appealing due to its statistical exactness for dealing with two different probability distributions under the absolute continuity condition \citep{sugiyama2007covariate}. Specifically, we define the \textbf{target group accuracy} of a group $\gG_n^{(t)}$ with the true IW $w^*(x) := \tfrac{p_T(x)}{p_S(x)}$ as 
\begin{equation}
    \alpha_T(\gG_n^{(t)}; w^*) = \E_{p_S}\left[w^*(X) \textbf{1}(Y = \hat{Y}) | \gG_n^{(t)}\right] \tfrac{P(X_S \in \gG_n^{(t)})}{P(X_T \in \gG_n^{(t)})} \label{eq:target_acc_estim}
\end{equation}
where $X_S$ and $X_T$ are random variables having densities $p_S$ and $p_T$, respectively. We denote $\hat{\alpha}_T(\gG_n^{(t)}; w^*)$ to be the expectation with respect to the empirical measure. We also define the \textbf{source group accuracy} of $\gG_n^{(t)}$ as 
\begin{equation}
    \alpha_S(\gG_n^{(t)}; w^*) = \E_{p_T}\left[ \tfrac{\textbf{1}(Y(X) = \hat{Y}(X))}{w^*(X)} | \gG_n^{(t)} \right]  \tfrac{P(X_T \in \gG_n^{(t)})}{P(X_S \in \gG_n^{(t)})}. \label{eq:source_acc_estim}
\end{equation}

Given $\alpha_T(\gG_n^{(t)}; w^*)$, the group accuracy estimation problem can be reduced to the importance weight estimation problem; that is, finding $\tilde{w}$ such that $\alpha_T(\gG_n^{(t)}; \tilde{w}) \approx \alpha_T(\gG_n^{(t)}; w^*)$. A typical approach for solving this estimation problem is to accurately approximate IW, i.e., $\tilde{w} \approx w^*$, which is challenging in high-dimensional spaces. In this work, we propose an optimization-based approach that minimizes the group accuracy estimation error during the IW estimation process, circumventing the difficulty of directly estimating $w^*$.

\subsection{Motivation for an Optimization-Based Approach}
Our idea for accurately estimating the ``target'' group accuracy with IW estimator $\tilde{w}$ is to define two estimators for the ``source'' group accuracy defined in \eqref{eq:source_acc_estim}, with one estimator dependent on $\tilde{w}$, and to encourage the two estimators to agree with each other. This approach can be validated because the target accuracy estimation error of $\tilde{w}$ can be upper bounded by its source accuracy estimation error; that is,
\begin{multline} \label{eq:target_acc_estim_bound}
    |\alpha_T(\gG_n^{(t)}; w^*) - \alpha_T(\gG_n^{(t)}; \tilde{w})| \leq \\
    \tilde{w}_n^{(ub)} \cdot |\alpha_S(\gG_n^{(t)}; w^*) - \alpha_S(\gG_n^{(t)}; \tilde{w})| (\tfrac{P(X_T \in \gG_n^{(t)})}{P(X_S \in \gG_n^{(t)})} )^2
\end{multline}
where $\tilde{w}_n^{(ub)} = \sup_{x \in Supp(p_T(\cdot | \gG_n^{(t)}))} \tilde{w}(x) $ and the bound is tight when $\tilde{w}(x) = \tilde{w}_n^{(ub)}$ for all $x \in Supp(p_T(\cdot | \gG_n^{(t)}))$.

Based on the fact that labeled samples are available in the source domain, we define a first estimator of $\alpha_S(\gG_n^{(t)}; w^*)$ with the simple Monte-Carlo estimation by
\begin{equation}
    \hat{\alpha}_S^{(MC)}(\gG_n^{(t)}) = \hat{\E}_{p_S}[\textbf{1}(Y= \hat{Y}) | \gG_n^{(t)}].
\end{equation}
We note that $\hat{\alpha}_S^{(MC)}(\gG_n^{(t)})$ serves as a guide for the agreement between two estimators because it accurately estimates $\alpha_S(\gG_n^{(t)}; w^*)$ with a small error of $\gO(1 / \sqrt{|\gG_n^{(t)}(\gD_S)|})$ where $\gG_n^{(t)}(\gD_S) := \{(x_k, y_k) \in \gD_S : x_k \in \gG_n^{(t)} \}$.

Based on the fact that input features are available both in source and target domains, we define a second estimator of $\alpha_S(\gG_n^{(t)}; w^*)$ with importance weighting. Specifically, by assuming $\E_{p_{T_{Y|x}}}[\textbf{1}(Y(x) = \hat{Y}(x))] = \alpha_T(\gG_n^{(t)}; w^*)$ for all $x \in \gG_n^{(t)}$ and replacing $w^*$ by $\tilde{w}$ in \eqref{eq:source_acc_estim}, we obtain the second estimator as a function of $\tilde{w}$, which is given by
\begin{multline} \label{eq:IW_souce_acc}
    \hat{\alpha}_S^{(IW)}(\gG_n^{(t)}; \tilde{w}) 
    := \tfrac{\hat{P}(X_T \in \gG_n^{(t)})}{\hat{P}(X_S \in \gG_n^{(t)})} \cdot \hat{\E}_{p_T}\left[ \tfrac{\hat{\alpha}_T(\gG_n^{(t)}; \tilde{w}) }{\tilde{w}(X)} | \gG_n^{(t)} \right] \\ 
    = \hat{\E}_{p_T}[ \tfrac{1}{\tilde{w}(X)} | \gG_n^{(t)} ] \hat{\E}_{p_S}[ \textbf{1}(Y = \hat{Y}) \tilde{w}(X) | \gG_n^{(t)} ] 
\end{multline}
where $\hat{\alpha}_T(\gG_n^{(t)}; \tilde{w})$ is an empirical estimate of the target accuracy defined in \eqref{eq:target_acc_estim}, $\hat{P}(X_T \in \gG_n^{(t)}) := \hat{\E}_{p_T}[\textbf{1}(X \in \gG_n^{(t)})] $, and $\hat{P}(X_S \in \gG_n^{(t)}) := \hat{\E}_{p_S}[\textbf{1}(X \in \gG_n^{(t)})] $.

Crucially, two estimators, $\hat{\alpha}_S^{(MC)}(\gG_n^{(t)})$ and $\hat{\alpha}_S^{(IW)}(\gG_n^{(t)}; \tilde{w})$, can be similar to each other \textit{if the target group accuracy estimation with $\tilde{w}$ is accurate} (cf. \eqref{eq:IW_souce_acc}); that is, $\hat{\alpha}_T(\gG_n^{(t)}; \tilde{w}) \approx \hat{\alpha}_T(\gG_n^{(t)}; w^*)$. Therefore, by encouraging consistency between two estimators via solving the optimization problem developed in Section \ref{subsec:tract_opt}, IW-GAE can accurately estimate the group accuracy not only in the source domain but also in the target domain. This conceptual attractiveness is empirically verified in Figure \ref{fig:aa}, which compares group accuracy estimation errors in the source and target domains from 720 IWs found by IW-GAE. Specifically, we found that under the optimal IW found by IW-GAE, group accuracy estimation errors in the source and target domains, $|\alpha_S(\gG_n^{(t)}; w^*) - \alpha_S(\gG_n^{(t)}; \tilde{w})|$ and $|\alpha_T(\gG_n^{(t)}; w^*) - \alpha_T(\gG_n^{(t)}; \tilde{w})|$ in \eqref{eq:target_acc_estim_bound}, are strongly correlated with a high Pearson correlation coefficient of 0.8.

\subsection{Formulating an Optimization Problem} \label{subsec:tract_opt}
In this section, we aim to solve an optimization problem such that $\min_{t, \tilde{w}} (\hat{\alpha}_S^{(IW)}(\gG_n^{(t)}; \tilde{w})- \hat{\alpha}_S^{(MC)}(\gG_n^{(t)}))^2$. Unfortunately, $\hat{\alpha}_S^{(IW)}(\gG_n^{(t)}; \tilde{w})$ in \eqref{eq:IW_souce_acc} is non-convex with respect to $\tilde{w}$ and non-smooth with respect to $t$, which is in general not effectively solvable with optimization methods \citep{jain2017non}. Further, directly solving an optimization problem in the function space of $\tilde{w}$ or optimizing over IW values for each $x \in \gX$ would be computationally demanding. Therefore, we introduce the following techniques to formulate the optimization problem in a tractable way. 

\textbf{Relaxed reformulation } 
We separately estimate IWs for the source and target domains, denoted as $\tilde{w}^{(S)}$ and $\tilde{w}^{(T)}$, and then encourage their agreement through constraints. As a result, the estimator in \eqref{eq:IW_souce_acc} becomes $\hat{\alpha}_S^{(IW)}(\gG_n^{(t)}; \tilde{w}^{(S)}, \tilde{w}^{(T)}) = \hat{\E}_{p_T}[ \tfrac{1}{\tilde{w}^{(T)}(X)} | \gG_n^{(t)} ] \hat{\E}_{p_S}[ \textbf{1}(Y = \hat{Y}) \tilde{w}^{(S)}(X) | \gG_n^{(t)} ]$, which is coordinatewise convex.

\textbf{Discretize $\gT$ } 
We use a discrete set $\gT : = \{t_1, t_2, \cdots, t_n\}$ based on the facts that a group separation is not sensitive to small changes in $t$ and the inner optimization is not smooth with respect to $t$. We remark that the inner optimization problem with respect to $\tilde{w}^{(S)}$ and $\tilde{w}^{(T)}$ is readily solvable, so the discrete optimization over $\gT$ can be performed without much computational overhead.

\textbf{Binned IWs } We approximate $\tilde{w}^{(S)}$ and $\tilde{w}^{(T)}$ by the binned IWs. Specifically, $\gX$ is partitioned into $B$ number of bins: $\gX = \cup_{i=1}^{B} \gB_i$ where $\gB_i = \{x \in \gX | I^{(B)}(x) = i \}$ and $I^{(B)}: \gX \rightarrow [B]$. Then, we assign the same IW value to all samples in the same bin; that is, $\tilde{w}^{(S)}(x_S) = \tilde{w}_j^{(S)}$ for  $x_S \in \gB_j \cap \gD_S$ and $\tilde{w}^{(T)}(x_T) = \tilde{w}_j^{(T)}$ for  $x_T \in \gB_j \cap \gD_T$. In this way, the number of decision variables in the inner optimization problem is reduced to $2B$. Further, we incorporate the recently proposed confidence interval (CI) estimation method for the binned IWs \citep{park2021pac} into the constraints. We denote $\Phi_j$ be the CI of the true binned IW of $\gB_j$, which is obtained by applying the Clopper-Pearson CI \citep{clopper1934use} (cf. Appendix \ref{appx:cp_ci_explain}). We let $\tilde{\textbf{w}}^{(S)} = (\tilde{w}_1^{(S)}, \cdots, \tilde{w}_B^{(S)})$ and $\tilde{w}^{(S)}(x) = \tilde{w}_{I^{(B)}(x)}^{(S)}$. We also define $\tilde{\textbf{w}}^{(T)}$ and $\tilde{w}^{(T)}(x)$ in the same way.

Assembling the three techniques, we can effectively solve the group accuracy estimation problem by finding binned IWs $ w^\dagger(n; t^\dagger) \in \R^{2B}_+$ for $n \in [M]$ by solving the following nested optimization (see Algorithm \ref{alg:iw_gae} for pseudocode): $t^\dagger \in \argmin_{t \in \gT}  \sum_{n\in[M]}  ( \hat{\alpha}_S^{(MC)}(\gG_n^{(t)}) - \hat{\alpha}_S^{(IW)}(\gG_n^{(t)}; w^\dagger(n; t)) )^2$ where $w^\dagger(n; t) $ is a solution of 
\begin{align} \label{eq:opt_prob}
    \min_{\tilde{\textbf{w}}^{(S)}, \tilde{\textbf{w}}^{(T)}} &  \: ( \hat{\alpha}_S^{(MC)}(\gG_n^{(t)}) - \hat{\alpha}_S^{(IW)}(\gG_n^{(t)}; \tilde{\textbf{w}}^{(S)}, \tilde{\textbf{w}}^{(T)}) )^2 \\ 
    \text{s.t. } & \: \tilde{w}_i^{(S)} \in \Phi_i,  \quad \text{for } i \in [B]  \label{eq:ci_const1} \\
    & \: \tilde{w}_i^{(T)} \in \Phi_i, \quad \text{for } i \in [B] \label{eq:ci_const2} \\
    & \: \parallel \tilde{w}_i^{(T)} - \tilde{w}_i^{(S)} \parallel_2^2 \leq \delta_{tol}, \quad \text{for } i \in [B] \label{eq:relax_const} \\ 
    & \: \left| \hat{\E}_{p_S}[\tilde{w}^{(S)}(X)| \gG_n^{(t)}] - \tfrac{\hat{P}(X_T \in \gG_n^{(t)})}{\hat{P}(X_S \in \gG_n^{(t)})} \right| \leq \delta_{pr} \label{eq:const1} \\ 
    & \: \left| \hat{\E}_{p_T}[\tfrac{1}{\tilde{w}^{(T)}(X)}| \gG_n^{(t)}] - \tfrac{\hat{P}(X_S \in \gG_n^{(t)})}{\hat{P}(X_T \in \gG_n^{(t)})} \right| \leq \delta_{pr}  \label{eq:const2}
\end{align}
where $\delta_{tol}$ and $\delta_{pr}$ are small constants. Box constraints \eqref{eq:ci_const1} and \eqref{eq:ci_const2} ensure that the obtained solution is in the CI, which bounds the estimation error of $\tilde{w}_i^{(S)}$ and $\tilde{w}_i^{(T)}$ by $|\Phi_i|$ and guarantees their asymptotic convergences to the true binned IW \citep{thulin2014cost}. This can also bound the target group  accuracy estimation error as $|\alpha_T(\gG_n^{(t)}; w^*) - \alpha_T(\gG_n^{(t)}; \tilde{\textbf{w}}^{(S)})| \leq \max_{b \in [B]}|\Phi_b| P(X_S \in \gG_n^{(t)}) / P(X_T \in \gG_n^{(t)}) $. Constraint \eqref{eq:relax_const} corresponds to the relaxation for removing non-convexity of the original objective, and $\delta_{tol} = 0$ recovers the original objective. Constraints \eqref{eq:const1} and \eqref{eq:const2} are based on the equalities that the true IW $w^*(\cdot)$ satisfies: $\E_{p_S}[w^*(X)| X \in \gG_n^{(t)}] = \tfrac{P(X_T \in \gG_n^{(t)})}{P(X_S \in \gG_n^{(t)})}$ and $\E_{p_T}[1 / w^*(X)| X \in \gG_n^{(t)}] = \tfrac{P(X_S \in \gG_n^{(t)})}{P(X_T \in \gG_n^{(t)})}$. After solving the optimization, we use the first $B$ elements of $w^\dagger(n; t^\dagger)$ that correspond to the optimal $\tilde{\textbf{w}}^{(S)}$ for estimating group accuracy of $\gG_n$, which is denoted by $w^\dagger(n)$.

\subsection{Analyzing the Optimization Problem}

\begin{table*}[]
\vskip -0.1in
\caption{Model calibration benchmark results of MDD (OfficeHome) and CDAN (DomainNet and VisDa-2017). 
The numbers indicate the mean ECE across ten repetitions with boldface for the minimum mean ECE. 
Due to space limitations, we present the first six domain pairs of OfficeHome and DomainNet in the main body and the rest of them in Tables \ref{table:full_office_home_ece} and \ref{table:full_domain_net}, respectively. However, we report average performance among all pairs in Avg*. Oracle is obtained by applying TS with labeled test samples in the target domain. }
\vskip 0.1in
\scalebox{0.79}{
\begin{tabular}{l|rrrrrr|r||rrrrrr|r||r}
\hline 
& \multicolumn{6}{c}{\textbf{OfficeHome}} & & \multicolumn{6}{c}{\textbf{DomainNet}} & & \textbf{VisDa-2017} \\
\hline
Method  & Ar-Cl & Ar-Pr & Ar-Rw & Cl-Ar & Cl-Pr & Cl-Rw & Avg*  & Cl-Pt &    Cl-Rw &   Cl-Sk &   Pt-Cl &   Pt-Rw &   Pt-Sk &  Avg* & Sim-Rw \\ \hline
Vanilla & 40.61 &        25.62 &         15.56 &         33.83 &         25.34 &         24.75 &   24.37 & 13.23 &        6.36 &          12.92 &         9.75 &          6.35 &          15.56 &  10.06 & 21.63 \\
TS & 35.86 &        22.84 &         10.60 &          28.24 &         20.74 &         20.06 &   24.01 & 12.95 &        \textbf{5.95} &          13.32 &         6.40 &   3.90 &          11.07 &   9.22 & 22.42 \\
CPCS  & 22.93 &        22.07 &         10.19 &         26.88 &         18.36 &         14.05 &  19.79 & \textbf{5.64} &         21.90 &          7.70 &   \textbf{5.14} &          7.72 &          7.90 &          9.60 & 22.42 \\
IW-TS & 32.63 &        22.90 &          11.27 &         28.05 &         19.65 &         18.67 & 23.26 & 16.76 &        16.70 &          12.53 &         5.29 &          7.84 &          4.34 &   10.49 & 22.19 \\
TransCal & 33.57 &        20.27 &         \textbf{8.88} &          26.36 &         18.81 &         18.42 & 20.84 & 18.51 &        29.63 &         20.92 &         23.02 &         31.83 &         17.58  &         25.39 & 18.79 \\
IW-GAE & \textbf{12.78} &        \textbf{4.70} &   12.93 &         \textbf{7.52} &          \textbf{4.42} &          \textbf{4.11} & \textbf{8.93} & 6.06 &         8.15 &          \textbf{5.38} &          7.45 &   \textbf{3.89} &          \textbf{3.94} &    \textbf{6.32} & \textbf{14.70} \\ 
\hline
Oracle & 10.45 &        10.72 &         6.47 &          8.10 &   7.62 &          6.55 &  8.42 & 4.55 &         2.78 &          4.01 &          3.10 &   3.72 &          2.72 &   3.02 & 5.48 \\
\hline
\end{tabular}}
\label{table:model_calib}
\vskip -0.05in
\end{table*}

The optimization problem in \eqref{eq:opt_prob}-\eqref{eq:const2} aims to estimate the truncated IW $w^*(x | \gG_n^{(t)}) := \frac{p_T(x | \gG_n^{(t)})}{p_S(x | \gG_n^{(t)})}$ for each $\gG_n^{(t)}$ that can induce an accurate source group accuracy estimator. 
However, the astute reader might notice that the objective in \eqref{eq:opt_prob} does not measure the source group accuracy estimation error. In the following proposition, we show that solving the optimization problem minimizes the upper bound of the source group accuracy estimation error, thereby the target group accuracy estimation error due to \eqref{eq:target_acc_estim_bound}.  
\begin{proposition} \label{proposition:closeness_opt}
    Let $w^\dagger(n)$ be a solution to the nested optimization problem with $\delta_{tol} = 0$ and $\delta_{pr} = 0$. 
    Let $\epsilon_{opt}(w^\dagger(n)) := ( \hat{\alpha}_S^{(MC)}(\gG_n^{(t)}) - \hat{\alpha}_S^{(IW)}(\gG_n^{(t)}; w^\dagger(n)) )^2$ be the objective value. 
    For $\tilde{\delta} > 0$, the following inequality holds with probability at least $1 - \tilde{\delta}$:
    \begin{align} 
        & |\alpha_{S}(\gG_n^{(t)}; w^*) - \alpha_{S}(\gG_n^{(t)}; w^\dagger(n))| \\
        & \leq \epsilon_{opt}(w^\dagger(n)) + \epsilon_{stat} + IdentBias(w^\dagger(n); \gG_n^{(t)}) \label{eq:tightning_obj}
    \end{align}
    where $\epsilon_{stat} \in \gO(\log(1 / {\tilde{\delta}}) / \sqrt{|\gG_n^{(t)}(\gD_S)|})$ and $IdentBias(w^\dagger(n); \gG_n^{(t)}) = \tfrac{P(X_T \in \gG_n^{(t)})}{2P(X_S \in \gG_n^{(t)})} ( \E_{p_T}[(\textbf{1}(Y = \hat{Y}) - \hat{\alpha}_T(\gG_n^{(t)}; w^\dagger(n)))^2 | \gG_n^{(t)} ] + 1 / \ubar{w}^\dagger(n)^2 ) $ for $\ubar{w}^\dagger(n) := \min_{i \in [B]} \{ w^\dagger_i(n) \}$ . 
\end{proposition}
The proof is based on the Cauchy-Schwarz inequality, which is provided in Appendix \ref{appx:proof_prop_ub_source_acc}. Proposition~\ref{proposition:closeness_opt} shows that we can reduce the source accuracy estimation error by reducing $\epsilon_{opt}(w^\dagger(n))$ by solving the optimization problem. Here, we note that a large value of $\epsilon_{stat} +  IdentBias(w^\dagger(n); \gG_n^{(t)}) $ or a looseness of \eqref{eq:tightning_obj} could significantly decrease the effectiveness of IW-GAE. However, in the empirical analyses presented in Figures \ref{fig:bb} and \ref{fig:cc}, it turns out that reducing $\epsilon_{opt}(w^\dagger(n))$ can effectively reduce the group accuracy estimation in both source and target domains. Finally, we note that $IdentBias(w^\dagger(n); \gG_n)$ can be reduced by decreasing the variance of the correctness within the group, which advocates our group construction with the maximum value of the softmax output (cf. Proposition \ref{proposition:bias_var_decomp}).

\section{Experiments}
In this section, we extensively evaluate IW-GAE on model calibration and selection tasks. Since both tasks are based on UDA classification tasks, we first provide the common setup and task-specific setup such as the baselines and evaluation metrics in the corresponding sections. 

\paragraph{Datasets } We use \textbf{\underline{OfficeHome}} \citep{venkateswara2017deep} containing around 15,000 images of 65 categories from four domains (art, clipart, product, real-world), \textbf{\underline{VisDA-2017}} \citep{peng2017visda} containing around 280,000 images of 12 categories from two domains (real and synthetic images), and \textbf{\underline{DomainNet}} \citep{peng2019moment} containing around 570,000 images of 345 categories from six domain pairs (clipart, real, sketch, infograph, painting, quickdraw). 

\textbf{Base models } We consider maximum mean discrepancy (MDD; \citep{zhang2019bridging}), conditional domain adversarial network (CDAN; \citep{long2018conditional}), and maximum classifier discrepancy (MCD; \citep{saito2018maximum}) with ResNet-50 \citep{he2016deep} as the backbone neural network, which are the most popular high-performing UDA methods. Details about the training configurations are given in Appendix~\ref{appx:add_exp_details}.

\textbf{IW-GAE implementation details } 
We solve the optimization problem in \eqref{eq:opt_prob}-\eqref{eq:const2} by sequential least square programming \citep{kraft1988software} because it is a constrained nonlinear optimization problem with box constraints. Also, we set the number of groups $M=10$ and the number of bins $B=10$ for all experiments, which are taken from the standard range $[10,20]$ used for binning samples based on summary statistics \citep{guo2017calibration, park2021pac}. Finally, we set $\gT = \{0.85, 0.90, 0.95, 1.00, 1.05, 1.10 \}$. We provide further details in Appendix~\ref{appx:add_exp_details}.

\subsection{Model Calibration Performance}
\textbf{Setup \& Metric }
In this experiment, our goal is to match the confidence of a prediction to its expected accuracy in the target domain. Following the standard \citep{guo2017calibration,park2020calibrated,wang2020transferable}, we use expected calibration error (ECE) on the test dataset as a measure of calibration performance. The ECE measures the average absolute difference between the confidence and accuracy of binned groups, which is defined as
\begin{equation}
    ECE(\gD_T) =  \sum_{n\in[E]} \tfrac{|\gM_n|}{|\gD_T|} |\hat{\text{Acc}}(\gM_n) - \hat{\text{Conf}}(\gM_n)|
\end{equation}
where $\gM_n := \{x_i \in \gD_T | \tfrac{n-1}{E} \leq m(x_i; 1) < \tfrac{n}{E} \}$, $\hat{\text{Acc}}(\gM_n)$ is the average accuracy in $\gM_n$, and $\hat{\text{Conf}}(\gM_n)$ is the average confidence in $\gM_n$. For IW-GAE, $\hat{\text{Conf}}(\gM_n) = \tfrac{1}{|\gM_n|}\sum_{x \in \gM_n} \hat{\alpha}_T(\gG^{(t^\dagger)}_{I^{(g)}(x)}; w^\dagger(I^{(g)}(x)))$, which is the average of group accuracy estimations that each $x \in \gM_n$ belongs to. We use $E = 15$ following the standard value \citep{guo2017calibration, wang2020transferable}.

\paragraph{Baselines }
We consider the following five different baselines: 
The \textit{vanilla} method uses a maximum value of the softmax output as the confidence of the prediction. 
We also consider temperature scaling-based methods that adjust the temperature parameter by maximizing the following calibration measures: 
\textit{Temperature scaling (TS)} \citep{guo2017calibration}: the log-likelihood on the source validation dataset; 
\textit{IW temperature scaling (IW-TS)}: the log-likelihood on the importance weighted source validation dataset; 
\textit{Calibrated prediction with covariate shift (CPCS)}: the Brier score \citep{brier1950verification} on the importance weighted source validation dataset; 
\textit{TransCal} \citep{wang2020transferable}: the ECE on the importance weighted source validation dataset with a bias and variance reduction technique. These methods also use a maximum value of the (temperature-scaled) softmax output as the confidence. 
For methods with the IW, we use a logistic regression-based IW estimator as in \citet{wang2020transferable} (cf. Appendix \ref{appx:iw_estim_logistic}).

\begin{table*}[]
\vskip -0.1in
\caption{Checkpoint selection benchmark results of MDD with ResNet-50 on OfficeHome. The numbers indicate the mean test accuracy of selected model across ten repetitions with boldface for the maximum mean test accuracy. Due to space limitations, we present the first six domain pairs in the main body and the rest of them in Tables \ref{table:full_hyperparam} and \ref{table:full_check_pt}. However, we report average performance among all pairs in Avg*. Here, we also present two best methods among the target-only validation methods and the rest of them in Tables \ref{table:full_hyperparam} and \ref{table:full_check_pt}. Lower bound and Oracle indicate the accuracy of the models with the worst and best test accuracy, respectively.} 
\vskip 0.1in
\scalebox{0.85}{
\begin{tabular}{l|rrrrrr|r||rrrrrr|r}
\hline
 & \multicolumn{6}{c}{\textbf{Hyperparameter Selection}}  & & \multicolumn{6}{c}{\textbf{Checkpoint Selection}} \\
\hline
Method  & Ar-Cl & Ar-Pr & Ar-Rw & Cl-Ar & Cl-Pr & Cl-Rw & Avg*  & Ar-Cl & Ar-Pr & Ar-Rw & Cl-Ar & Cl-Pr & Cl-Rw & Avg* \\ \hline
Vanilla & 53.31 &        \textbf{70.96} &         77.44 &         59.70 &          65.17 &         69.96 & 65.45 & 47.22 &        74.14 &         77.76 &         61.85 &         70.96 &         71.59 &         67.47 \\

IWCV & 53.24 &        69.61 &         72.50 &          59.70 &          65.17 &         67.50 & 65.18 & {54.46} &        {74.22} &         72.27 &         61.48 &         70.49 &         70.62  &         67.48 \\
DEV & 53.31 &        70.72 &         77.44 &         59.79 &         67.99 &         69.96 & 66.00 & 54.04 &        73.94 &         78.16 &         61.52 &         63.19 &         70.70  & 67.39 \\ 
InfoMax & \textbf{54.34}     & \textbf{70.96}      & 77.53      & \textbf{61.48}      & \textbf{69.93}      & \textbf{71.06} & 67.79 & 54.32 & \textbf{74.72} & 77.90 & \textbf{62.79} & 71.03 & 71.47 & 68.38 \\ 
TransScore & \textbf{54.34}      & \textbf{70.96}      & 77.53      & \textbf{61.48}      & \textbf{69.93}      & \textbf{71.06}  & 67.87 & \textbf{54.79} & 74.14 & 77.77 & 61.76 & 70.97 & 71.48 & 68.38 \\ 
IW-GAE & \textbf{54.34} &        \textbf{70.96} &         \textbf{78.47} &         \textbf{61.48} &         \textbf{69.93} &         \textbf{71.06} & \textbf{67.95} & 54.32 &        73.98 &         \textbf{78.51} &         {61.96} &         \textbf{71.25} &         \textbf{71.70} &                \textbf{68.48} \\ \hline
Lower bound  & 52.51 &        69.27 &         72.50 &          59.70 &          65.17 &         67.50 &  64.10 & 41.90 &         64.88 &         72.27 &         52.00 &          58.48 &         62.13 &        58.21 \\   \hline
Oracle & 54.34 &        70.96 &         78.47 &         61.48 &         69.93 &         71.06 & 68.01 & 54.80 &         74.79 &         78.61 &         62.79 &         71.59 &         72.18 &  68.95 \\ \hline
\end{tabular}}
    \vskip -0.1in
\label{table:model_selec_table}
\end{table*}

\begin{table}[h!]
\vskip -0.1in
\caption{Additional model calibration benchmark with post-hoc calibration methods under unknown distribution shifts. Due to space limitations, we present the first six domain pairs in the main body and the rest of them in Table \ref{table:full_office_home_ece}. }
\vskip 0.1in
\scalebox{0.79}{
\begin{tabular}{l|rrrrrr|r}
\hline
Method  & Ar-Cl & Ar-Pr & Ar-Rw & Cl-Ar & Cl-Pr & Cl-Rw & Avg* \\ \hline
Vanilla  & 40.61 &        25.62 &         15.56 &         33.83 &         25.34 &         24.75 &        27.37 \\
PTS & 31.91 & 24.36 & 10.65 & 22.81 & 20.42 & 15.92  & 21.91 \\
AvUTS & 29.59 & 25.55 & 10.40 & 31.81 & 26.06 & 26.15  & 28.17 \\
TransCal   & 33.57 &        20.27 &         8.88 &          26.36 &         18.81 &         18.42  &          20.84 \\
IW-GAE  & {12.78} &        {4.70} &   12.93 &         {7.52} &          {4.42} &          {4.11} &  {8.93} \\ \hline
\end{tabular}}
\label{table:add_posthoc}
\vskip -0.2in
\end{table}

\paragraph{Results } As shown in Table~\ref{table:model_calib}, IW-GAE achieves the best average ECEs across different base models and datasets. 
Specifically, IW-GAE outperforms state-of-the-art performances by 53\% on OfficeHome, 31\% on DomainNet, and 21\% on VisDa-2017.
Further, in additional experiments with different base methods (CDAN and MCD) on OfficeHome, IW-GAE consistently outperforms state-of-the-art performances by 2\% and 5\%, respectively (cf. Tables \ref{table:add_cdan_ece} and \ref{table:add_mcd_ece}). 
Given that the second best model varies for a different dataset and a different base model, we believe that the consistent improvements by IW-GAE indicate its significant robustness compared to the baselines.

In Table \ref{table:add_posthoc}, we also compare IW-GAE with two recent post-hoc calibration methods, called PTS \citep{tomani2022parameterized} and AvUTS \citep{krishnan2020improving}, which are designed to perform model calibration without using target domain samples. As post-hoc calibration methods under general distribution shifts, PTS and AvUTS show compatible results with IW-GAE and outperform some baselines in some domain pairs, i.e., certain types of distribution shifts. However, they cannot achieve better average performances than baselines or IW-GAE that explicitly consider the target distribution shift through importance weighting. The results show the advantage of explicitly using the information about the distribution shifts for the model calibration task in UDA.

\subsection{Model Selection Performance}
\textbf{Setup \& Metric }
In this experiment, we perform two important model selection tasks of choosing the best checkpoint and the best hyperparameter.
Specifically, for the checkpoint selection, we train MDD on the OfficeHome dataset for 30 epochs and save the checkpoint at the end of each epoch. For the hyperparameter selection, we repeat training the MDD method by changing its key hyperparameter of margin coefficient from 1 to 8 (the default value is 4). Given a set of models from different checkpoints or different hyperparameters, we choose the best model based on a model selection criterion (such as IW-GAE or other baselines). Specifically, for IW-GAE, we choose the model with the maximum value of the mean group accuracy estimations, which is computed by $\tfrac{1}{|\gD_T|}\sum_{x \in \gD_T} \hat{\alpha}_T(\gG^{(t^\dagger)}_{I^{(g)}(x)}; w^\dagger(I^{(g)}(x)))$. Then, we compare the test target accuracy of the chosen models under different model selection methods.

\paragraph{Baselines }
We consider the following baselines that evaluate the model's performance in terms of the following criterion:
\textit{Vanilla}: the minimum classification error on the source validation dataset;
\textit{Importance weighted cross validation (IWCV)} \citep{sugiyama2007covariate}: the  minimum importance-weighted classification error on the source validation dataset;
\textit{Deep embedded validation (DEV))} \citep{you2019towards}: the minimum deep embedded validation risk on the source validation dataset;
\textit{Target-only validation methods}: the maximum value of some pre-defined measures, e.g., the average negative entropy of predictions, on the unlabeled target validation dataset.
Again, we use a logistic regression-based IW estimator for methods with the IW (IWCV and DEV).
For the target-only validation methods, we consider InfoMax \citep{shi2012infomax}, Corr-C \citep{tu2022assessing}, SND \citep{saito2021tune}, MixVal \citep{hu2024mixed}, and TransScore \citep{yang2023can}.

\paragraph{Results }
Table~\ref{table:model_selec_table} shows that model selection with IW-GAE achieves the best average accuracy among IW-based model selection methods, improving state-of-the-art by 9\% in the checkpoint selection 18\% in the hyperparameter selection in terms of the relative scale of lower and upper bounds of accuracy. Note that IWCV does not improve the vanilla method on average, which could be due to the inaccurate IW estimation by the logistic regression-based method. In this sense, IW-GAE has the advantage of depending less on the performance of the IW estimator since the estimated value is used to construct bins for the CI, and then the exact value is found by solving the separate optimization problem.

In addition, we observe that all target-only validation methods except SND outperform the IW-based baselines (cf. Table \ref{table:model_selec_table}). The results are consistent with the recent empirical observations that target-only validation methods are more favorable than IW-based methods for the model selection task in UDA \citep{saito2021tune, hu2024mixed}. However, notably, they underperform IW-GAE in both checkpoint and hyperparameter selection tasks, which strongly supports the practical advantage of IW-GAE as a model selection method. We believe that the impressive empirical performance of IW-GAE could bring more attention to the IW-based model selection techniques in the machine learning community, which is a principled statistical method for the model selection task under distribution shifts but has been considered impractical and less effective.

\subsection{Ablation Study} \label{sec:ablation}
We conduct an ablation study by performing the model calibration task without key components of IW-GAE.  

\textbf{Group construction with the softmax output }
We first examine the effectiveness of group construction based on the maximum value of the softmax output by examining a group construction function based on IW. 
In Table \ref{table:ablation}, we can see that grouping by the IW significantly reduces the performance of IW-GAE due to a large variance of prediction accuracy within a group (cf. the case of group 2 in Figure~\ref{fig:figure1b}). 
Specifically, the large value of $Var(\textbf{1}(Y = \hat{Y}) | \gG_n^{(t)} ) $ increases the $IdentBias(w^\dagger(n); \gG_n^{(t)})$, which can loosen the upper bound of the source group accuracy estimation error in \eqref{eq:tightning_obj}. 
Therefore, it is important to construct groups so that each group has a low variance of the correctness of predictions, as our design developed in Section~\ref{sec:gr_assign}.

\begin{table}[]
\vskip -0.1in
\caption{Results of an ablation study with four randomly selected domain pairs in OfficeHome. The numbers indicate the mean ECE of MDD with ResNet-50.}
\vskip 0.1in
\scalebox{0.85}{
\begin{tabular}{l|rrrr | r}
\hline
Method  & Ar-Pr & Pr-Cl & Rw-Cl & Rw-Pr & Avg \\ 
\hline
Vanilla &  40.61   &  38.62    &   36.51  & 14.01     & 32.44 \\
CPCS & 22.07   & 29.20    &   26.54  &   11.14  & 22.24 \\
TransCal & 20.27 & 29.86    &  29.90   &    10.00 & 22.51 \\
\hline 
Grouping by IW   & 14.18 & 34.67 & 35.08 & 5.30 & 22.31 \\
W/O CI & 11.00 & 29.73 & 24.44 & 2.09 & 16.82 \\
IW-Mid & 31.62 & 30.35 & 26.32 & 10.60 & 24.70 \\
\hline
IW-GAE  & 4.70 & 17.49 & 9.52 & 8.14 & 9.97 \\
\hline
\end{tabular}}
\label{table:ablation}
\vskip -0.2in
\end{table}

\begin{figure}
    \vskip -0.05in
    \centering{
    \begin{subfigure}
     {\includegraphics[width=0.235\textwidth]{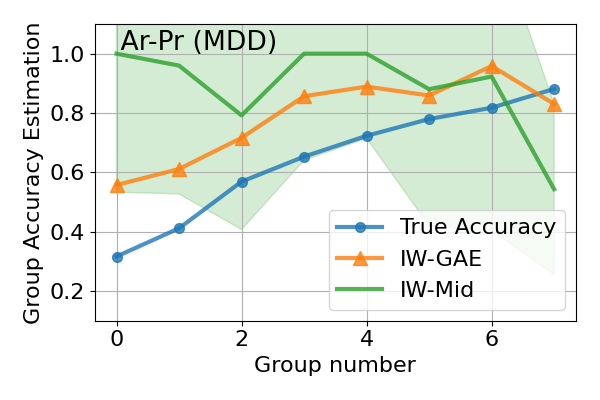}}
    \end{subfigure}
    \begin{subfigure}
     {\includegraphics[width=0.23\textwidth]{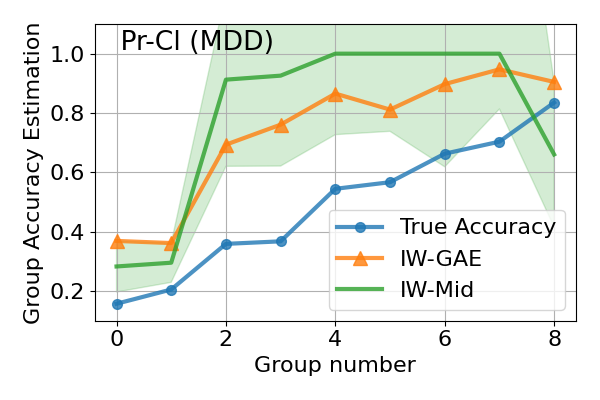}}
    \end{subfigure} 
    \\
    \vskip -0.2in
    \begin{subfigure}
     {\includegraphics[width=0.23\textwidth]{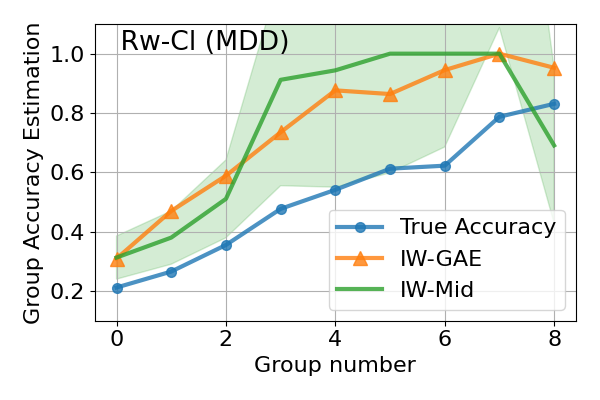}}
    \end{subfigure}
    \begin{subfigure}
     {\includegraphics[width=0.235\textwidth]{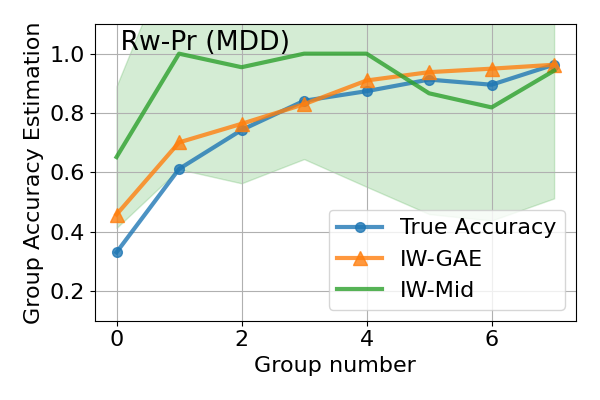}}
    \end{subfigure}
    }
    \vskip -0.2in
    \caption{True group accuracy and estimated group accuracy of IW-GAE and IW-Mid under MDD. The shaded areas represent possible group accuracy estimation with binned IWs in the CI. 
    See Figure \ref{fig:rc_curve} for visualization of all domain pairs.}
    \label{fig:quan_anal}
    \vskip -0.1in
\end{figure}

\textbf{Constraints from the CI estimation method }
Next, we examine the dependency of the effectiveness of IW-GAE on the CI estimation method \citep{park2021pac} by setting only minimum and maximum values of IWs (W/O CI); that is, $\Phi_i = [1/6, 6.0]$ for $i \in [M]$. In Table \ref{table:ablation}, we can see that IW-GAE outperforms strong baseline methods (CPCS and TransCal) even under this naive interval of IWs. However, the performance is reduced compared to the setting with the sophisticated CI estimator. In this regard, developing an optimization-based IW estimation method that works effectively without the CI estimator in the constraints could be an interesting future direction.

\textbf{IW optimization }
To further show the effectiveness of the optimization in IW-GAE, we also test the method of selecting the middle point in the CI proposed in \citet{park2021pac} as an IW estimator (IW-Mid), which originates herein. For IW-Mid, we perform both model calibration and selection tasks across different base models and datasets. Surprisingly, IW-Mid achieves the better average performances than other IW-based baselines in some benchmarks (cf. Tables \ref{table:full_office_home_ece}-\ref{table:full_check_pt}). However, its performance is worse and significantly unstable compared to IW-GAE, such as achieving the worse performance even than the vanilla method in experiments with CDAN (Table \ref{table:add_cdan_ece}) and MCD (Table \ref{table:add_mcd_ece}). This means that the CI estimation method does not effectively estimate the group accuracy without properly selecting the exact IW through our optimization method, which is consistent with the theoretical result in Proposition~\ref{proposition:closeness_opt}. The instability and inaccuracy of IW-Mid compared to IW-GAE also can be identified in the qualitative evaluation of their group accuracy estimations (cf. Figure \ref{fig:quan_anal}), which shows that the true accuracy is close to IW-Mid only in some cases.

\begin{figure}
    \centering
    \hfill
    \begin{subfigure}[]
     {\includegraphics[width=0.231\textwidth]{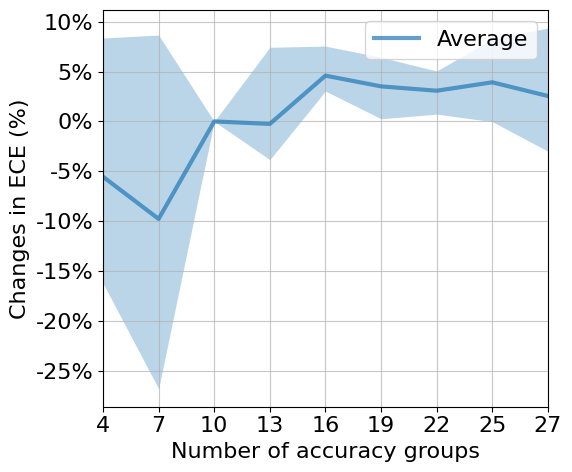}}
    \end{subfigure}
    \hfill
    \begin{subfigure}[]
     {\includegraphics[width=0.231\textwidth]{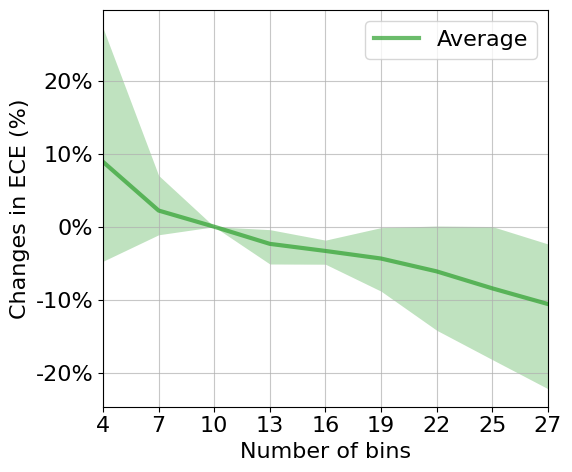}}
    \end{subfigure}
    \hfill
    \vskip -0.1in
    \caption{Sensitivity analysis with respect to $M$ (a) and $B$ (b) on four domain pairs (Ar-Pr, Pr-Cl, Rw-Cl, Rw-Pr) in OfficeHome. The shaded areas represent areas between the minimum and the maximum changes in ECE (lower is better). }
    \label{fig:senst_analy}
    \vskip -0.2in
\end{figure}

\subsection{Sensitivity Analysis} \label{sec:senst_analy}
We perform a sensitivity analysis with respect to key hyperparameters of the number of accuracy groups $M \in [4, 27] $ and the number of bins $B \in [4, 27]$ (the default value for both $M$ and $B$ is 10). In Figure \ref{fig:senst_analy}, note that the average performance changes under different hyperparameter values are somewhat stable; the average changes are within the range of 10\% for most cases, even though a large variance appears for extreme values such as $M = 4$ and $B = 27$. Therefore, the results show that IW-GAE would consistently outperform state-of-the-art methods under changes in $M$ and $B$ within their standard range $[10,20]$ since the best baseline (TransCal) achieves the mean ECE score about 40\% higher for the selected domain pairs (cf. Table \ref{table:model_calib}).

\section{Related Work}
\textbf{Model calibration in UDA} 
Although post-hoc calibration methods \citep{guo2017calibration} and Bayesian methods \citep{gal2016dropout, lakshminarayanan2017simple, sensoy2018evidential} have been achieving impressive calibration performances in the i.i.d. setting, it has been shown that most of the calibration improvement methods fall short under distribution shifts \citep{ovadia2019can} (see Appendix \ref{appx:literature_calib_iid} for more discussion). While there have been attempts to perform model calibration under unknown distribution shifts by simulating the distribution shifts \citep{salvador2021frustratingly}, designing a robust loss function that prevents extrapolations with high confidences \citep{krishnan2020improving, hebbalaguppe2022stitch, liu2022devil}, and learning an instant-wise temperature parameter \citep{tomani2022parameterized}, handling model calibration problems under general distribution shifts is challenging. However, the availability of unlabeled samples in the distribution shifted target domain relaxes the difficulty of model calibration in UDA. In particular, unlabeled samples in the target domain enable an IW formulation for the quantity of interests in the shifted domain. Therefore, the post-doc calibration methods (e.g., \citep{guo2017calibration}) can be applied by reweighting calibration measures such as the expected calibration error \citep{wang2020transferable} and the Brier score \citep{park2020calibrated} in the source dataset with an IW. However, it remains unclear how the IW estimation error impacts the calibration error. Our approach, by contrast, can directly minimize the calibration error during the IW estimation process.

\textbf{Model selection in UDA}
A standard procedure for model selection in the i.i.d. settings is the cross-validation, which enjoys statistical guarantees about bias and variance of model performance \citep{stone1977asymptotics, kohavi1995study, efron1997improvements}. However, in UDA, the distribution shifts violate assumptions for the statistical guarantees. Furthermore, in practice, the accuracy measured in one domain is significantly changed in the face of natural/adversarial distribution shifts \citep{goodfellow2014explaining, hendrycks2019benchmarking, ovadia2019can}. To tackle the distribution shift problem, importance weighted cross validation \citep{sugiyama2007covariate} applies importance sampling for obtaining an unbiased estimate of model performance in the distribution shifted target domain. Further, recent work in UDA controls variance of the importance-weighted cross validation with a control variate \citep{you2019towards}. These methods aim to accurately estimate the IW and then use an IW formula for the expected accuracy estimation. In this work, our method concerns the accuracy estimation error in the target domain during the process of IW estimation, which can potentially induce an IW estimation error but result in an accurate accuracy estimator. Finally, we remark a direction that aims to evaluate the model performance without source domain data based on neighborhood structure \citep{saito2021tune, hu2024mixed}, prediction uncertainty in the target domain \citep{musgrave2022benchmarking, tu2022assessing}, and newly developed metrics \citep{yang2023can}.

\section{Conclusion}
In this work, we address the model calibration and selection tasks in UDA by estimating group accuracy, which is accurately estimated by solving a novel optimization problem. Specifically, we define a Monte-Carlo estimator and an IW-based estimator of \textit{group accuracy in the source domain}. Then, we formulate an optimization problem that aims to find the IW making the two estimators close to each other. Crucially, the optimal IW provably leads to \textit{an accurate group accuracy estimator in the target domain}. Our method achieves the best performances in both model calibration and selection tasks in UDA across a wide range of benchmark problems. We believe that the impressive performance gains by our method show a promising future direction of the (importance-weighted) group accuracy estimation for addressing critical challenges in UDA.

\paragraph{Limitations and future directions}
We note that all IW-based methods (CPCS, IW-TS, TransCal, IW-GAE) fail to improve the standard method in the i.i.d. scenario in our experiments with pre-trained large-language models (XLM-R \citep{conneau2019unsupervised} and GPT-2 \citep{solaiman2019release}). We conjecture that these models are less subject to the distribution shifts due to massive amounts of training data that may include the target domain datasets, so applying the methods in the i.i.d. setting can work effectively. In this regard, we leave the following important research questions: “Are IW-based methods less effective under mild distribution shifts?” and “Can we develop methods effective for all levels of distribution shifts?”

\section*{Impact Statement}
In this work, we consider model calibration and model selection problems under distribution shifts, which hold great significance in practice, especially in safety-critical domains such as medical diagnosis and autonomous driving. Specifically, the well-calibrated models can properly require human intervention for uncertain predictions, preventing any catastrophic consequences from overconfident predictions in automated systems. In addition, an accurate model selection allows the deployment of high-performing models in the distribution-shifted environment. Further, precisely evaluating model performance enables practitioners to reject deploying the model if its estimated performance is below a certain acceptable level. As evidenced in our extensive evaluations, IW-GAE helps to obtain robust and trustworthy models equipped with these ideal properties in unsupervised domain adaptation settings. Also, the core idea behind IW-GAE is to introduce a notion of group accuracy and then estimate it by optimizing the importance weight, which does not depend on particular characteristics or attributes of datasets. Therefore, IW-GAE would not leverage any biases inherent in the datasets, which prevents our work from having potential negative societal consequences. To sum up, we believe in a positive broader impact of IW-GAE.

\section*{Acknowledgement}
We would like to thank Jihyeon Hyeong, Yuchen Lou, Jiezhong Wu, and anonymous reviewers for insightful discussions and helpful suggestions in writing the manuscript.

\bibliography{reference}
\bibliographystyle{icml2024}

%%%%%%%%%%%%%%%%%%%%%%%%%%%%%%%%%%%%%%%%%%%%%%%%%%%%%%%%%%%%%%%%%%%%%%%%%%%%%%%
%%%%%%%%%%%%%%%%%%%%%%%%%%%%%%%%%%%%%%%%%%%%%%%%%%%%%%%%%%%%%%%%%%%%%%%%%%%%%%%
% APPENDIX
%%%%%%%%%%%%%%%%%%%%%%%%%%%%%%%%%%%%%%%%%%%%%%%%%%%%%%%%%%%%%%%%%%%%%%%%%%%%%%%
%%%%%%%%%%%%%%%%%%%%%%%%%%%%%%%%%%%%%%%%%%%%%%%%%%%%%%%%%%%%%%%%%%%%%%%%%%%%%%%
\newpage
\appendix
\onecolumn

\setcounter{table}{0}
\renewcommand{\thetable}{A\arabic{table}}

\setcounter{figure}{0}
\renewcommand{\thefigure}{A\arabic{figure}}

\section{Proof of Claims} \label{section:appendix_proof}

\subsection{Proof of Proposition \ref{prop:suff_gae}} \label{appx:proof_suff}

\begin{proof}
The proof consists of three parts: 1) decomposition of the expected mean-square error of an estimator $g(x)$; 2) deriving MLEs of individual and group accuracies; 3) constructing a sufficient condition.

\textbf{1) Bias-variance decomposition of the expected mean-square error}
The expected mean-square error of an estimator $g(x)$ for $\beta(x)$ at $x_i \in \gG_n^{(t)}$ with respect to the realization of a label $y_i \sim Y|x$ can be decomposed by
\begin{equation} \label{eq:appx_bv_decomp}
    \E_{D} [(\hat{\beta}(x_i) - g(x_i) )^2] = 
    Var_D(g(x_i; D)) + ( Bias_{D}(g(x_i; D)) )^2 + \sigma^2_{x_i}
\end{equation}
where $Var_D(g(x_i; D)) := \E_D [(g(x_i; D) - \E_{D}[g(x_i;D)])^2]$ is the variance of the estimator and $Bias_D(g(x_i;D)) := \E_D[g(x_i;D)] - \beta(x)$ is the bias of the estimator.

\textbf{2) MLEs of individual and group accuracy estimators}
For an individual accuracy estimator $\hat{\beta}^{(id)}(x;D)$ that predicts an accuracy for each sample $x$ given $D$, an MLE estimator is $\hat{\beta}^{(id)}(x) = \hat{\beta}(x)$. This estimator is unbiased because $\E_{D}(\hat{\beta}^{(id)}(x;D)) = \beta(x)$ for each $x \in \gG_n^{(t)}$. Therefore, this estimator has the average of expected errors 
\begin{equation}
    \frac{1}{N_n} \sum_{k \in [N_n]} \E_{D} [(\hat{\beta}(x_k) - \hat{\beta}^{(id)}(x_k; D) )^2] = \bar{\sigma}^2 + \bar{\sigma}^2 \label{eq:err_mle_id}
\end{equation}
where $\bar{\sigma}^2 := \frac{1}{N_n}\sum_{i \in [N_n]} \sigma^2_{x_i}$.

For a group accuracy estimator $\hat{\beta}^{(gr)}(x;D)$ that predicts the same group accuracy estimate for all $x \in \gG_n^{(t)}$, an MLE estimator can be defined by $\hat{\beta}^{(gr)}(x; D) = \frac{1}{N_n} \sum_{i =1}^{N_n} \hat{\beta}(x_i)$, which is a biased estimator because $\E_{D}(\hat{\beta}^{(gr)}(x;D)) = \frac{1}{N_n} \sum_{i =1}^{N_n} \beta(x_i)$ for each $x \in \gG_n^{(t)}$. Therefore, this estimator has the average of expected errors
\begin{align}
    \frac{1}{N_n} \sum_{k \in [N_n]} \E_{D} [(\hat{\beta}(x_k) - \hat{\beta}^{(gr)}(x_k) )^2] 
    & = \frac{1}{N_n} \bar{\sigma}^2 + \frac{1}{N_n} \sum_{k \in [N_n]} ( \frac{1}{N_n} \sum_{i \in [N_n]} \beta(x_i) - \beta(x_k) )^2 + \bar{\sigma}^2 \\
    & = \frac{1}{N_n} \bar{\sigma}^2 + Var(\beta; D) + \bar{\sigma}^2 \label{eq:err_mle_gr}
\end{align}
where $Var(\beta; D)$ is the variance of the accuracy in group $\gG_n^{(t)}$.

\textbf{3) Sufficient condition}
Given \eqref{eq:err_mle_id} and \eqref{eq:err_mle_gr}, the Popoviciu's inequality \citep{popoviciu1965certaines} provides a sufficient condition for the group accuracy estimator $\hat{\beta}^{(gr)}$ to have a lower expected mean-squared error than the individual accuracy estimator $\hat{\beta}^{(id)}$ as follows:
\begin{equation}
    Var(\beta; D) \leq \frac{1}{4} ( \max_{x^\prime \in \gG_n^{(t)}} \beta(x^\prime) - \min_{x^\prime \in \gG_n^{(t)}} \beta(x^\prime)  )^2 \leq \frac{N_n - 1}{N_n} \bar{\sigma}^2 = \frac{N_n - 1}{N_n} (\frac{1}{N_n} \sum_{i \in [N_n]} \beta(x_i) (1-\beta(x_i)) )
\end{equation}
where the equality comes from $\bar{\sigma}^2 = \frac{1}{N_n}\sum_{i \in [N_n]} \sigma^2_{x_i}$ with $\sigma^2_{x_i}$ is the variance of the Bernoulli distribution with a parameter $\beta(x_i)$.
\end{proof}

\subsection{Sufficient Condition for \eqref{eq:stat_favorable}} \label{appx:suff_conf_stat_favorable}

\begin{figure}
    \centering{\includegraphics[width=0.3\textwidth]{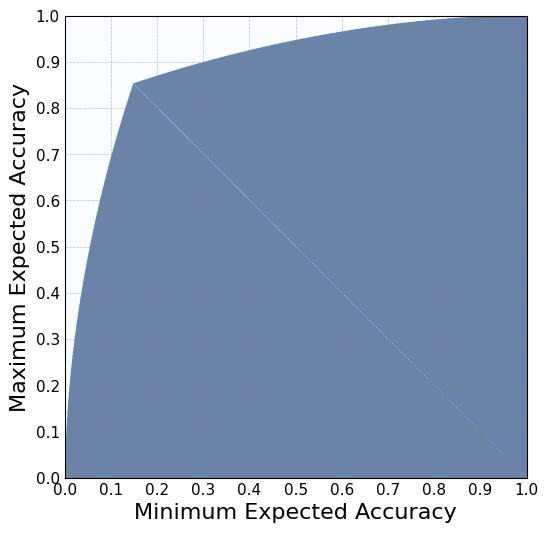}}
    \caption{The shaded area includes values of maximum and minimum expected accuracies within the group, which satisfy the sufficient condition for \eqref{eq:stat_favorable} when $N_n = 100$.}
    \label{fig:suff_conf_gad}
\end{figure}

To find the sufficient condition for \eqref{eq:stat_favorable}, i.e., $\tfrac{1}{4} ( \max_{x^\prime \in \gG_n} \beta(x^\prime) - \min_{x^\prime \in \gG_n} \beta(x^\prime)  )^2 \leq \tfrac{N_n - 1}{N_n} \bar{\sigma}^2$, note that $\bar{\sigma}^2 \geq \min_{x \in \gG_n} \sigma^2_{x} = \tilde{\beta} (1 - \tilde{\beta})$ where $\tilde{\beta} = \min\{\min_{x^\prime \in \gG_n} \beta(x^\prime), (1 - \max_{x^\prime \in \gG_n} \beta(x^\prime)) \}$. 
Therefore, the sufficient condition is $\max_{x^\prime \in \gG_n} \beta(x^\prime) - \min_{x^\prime \in \gG_n} \beta(x^\prime) \leq 2 \cdot \sqrt{\frac{N_n - 1}{N_n} \tilde{\beta} (1 - \tilde{\beta})}$, which is described in Figure \ref{fig:suff_conf_gad}. 

\subsection{Decomposition of the Expected Squared Calibration Error} 
\begin{proposition} \label{prop:brier_score_decomposition}
    The expected squared calibration error can be decomposed as the sum of the variance of the accuracies for samples in the same group and a squared group accuracy estimation error. That is, 
    \begin{equation}
        \E_{p_T}[(P(Y = \hat{Y}) - \hat{\alpha}_T(\gG^{(t)}_{I^{(g)}(X)}))^2] 
        = \sum_{n \in [M]} P(X_T \in \gG_n^{(t)}) [ Var(P(Y = \hat{Y})|\gG_n^{(t)}) + (\alpha_T(\gG^{(t)}_{n}) - \hat{\alpha}_T(\gG^{(t)}_{n}))^2 ]
    \end{equation}
    where $X_T$ is a random variable having a density of $p_T$.
\end{proposition}
\begin{proof}
    \begin{align}
        & \E_{p_T}[(P(Y = \hat{Y}) - \hat{\alpha}_T(\gG^{(t)}_{I^{(g)}(X)}))^2] 
        = \sum_{n \in [M]} P(X_T \in \gG^{(t)}_n) \E_{p_T}[(P(Y = \hat{Y}) - \hat{\alpha}_T(\gG^{(t)}_{I^{(g)}(X)}))^2| X \in \gG^{(t)}_n] \\
        & = \sum_{n \in [M]} P(X_T \in \gG^{(t)}_n) \E_{p_T}[(P(Y = \hat{Y}) - {\alpha}_T(\gG^{(t)}_{n}) + {\alpha}_T(\gG^{(t)}_{n})  - \hat{\alpha}_T(\gG^{(t)}_{n}))^2| X \in \gG^{(t)}_n] \\ 
        & = \sum_{n \in [M]} P(X_T \in \gG^{(t)}_n) \left( \E_{p_T}[(P(Y = \hat{Y}) - {\alpha}_T(\gG^{(t)}_{n}))^2| X \in \gG^{(t)}_n] 
        + ({\alpha}_T(\gG^{(t)}_{n})  - \hat{\alpha}_T(\gG^{(t)}_{n}))^2 \right) \label{eq:cross_term_brier} \\ 
        & = \sum_{n \in [M]} P(X_T \in \gG^{(t)}_n) [ Var(P(Y = \hat{Y})|\gG^{(t)}_n) + (\alpha_T(\gG^{(t)}_{n}) - \hat{\alpha}_T(\gG^{(t)}_{n}))^2 ]
    \end{align}
    where the equality \eqref{eq:cross_term_brier} holds due to $\E_{p_T}[ (P(Y(X) = \hat{Y}(X)) - \alpha_T(\gG^{(t)}_n))(\alpha_T(\gG^{(t)}_n) - \hat{\alpha}_T(\gG^{(t)}_n)) | X \in \gG^{(t)}_n] = 0 $.

\end{proof}

\subsection{Proof of Proposition \ref{proposition:closeness_opt}} \label{appx:proof_prop_ub_source_acc}

\begin{proof}
By applying triangle inequalities, we get the following inequality:
\begin{multline}
    |\alpha_{S}(\gG_n^{(t)}; w^*) - \alpha_{S}(\gG_n^{(t)}; w^\dagger(n))| 
    \leq 
    |\alpha_S(\gG_n^{(t)}; w^*) - \hat{\alpha}_S^{(MC)}(\gG_n^{(t)})|
    + |\hat{\alpha}_S^{(MC)}(\gG_n^{(t)}) - \hat{\alpha}^{(IW)}_{S}(\gG_n^{(t)}; w^\dagger(n))|
    \\
    + |\hat{\alpha}^{(IW)}_{S}(\gG_n^{(t)}; w^\dagger(n)) - {\alpha}^{(IW)}_{S}(\gG_n^{(t)}; w^\dagger(n))|
    +  |\alpha^{(IW)}_{S}(\gG_n^{(t)}; w^\dagger(n)) - \alpha_S(\gG_n^{(t)}; w^\dagger(n))|.
\end{multline}

Note that the first and third terms in the right hand side are coming from the Monte-Carlo approximation, so they can be bounded by $\gO(\log(1 / {\tilde{\delta}}) / |\gG_n^{(t)}(\gD_S)|)$ with probability at least $1 - {\tilde{\delta}}$ based on a concentration inequality such as the Hoeffding’s inequality. Also, the second term is bounded by the optimization error $\epsilon_{opt}(w^\dagger(n))$. Therefore, it is enough to analyze the fourth term. 

The fourth term is coming from the bias of  $\E_{T_{{Y|X}}}[\textbf{1}(Y(X) = \hat{Y}(X))] = \hat{\alpha}_T(\gG_n^{(t)}; w^\dagger(n))$, which we refer to as \textit{the bias of the identical accuracy assumption}. It can be bounded by
\begin{align}
    & |\alpha^{(IW)}_{S}(\gG_n^{(t)}; w^\dagger(n)) - \alpha_S(\gG_n^{(t)}; w^\dagger(n))| \\ 
    & = \frac{P(X_T \in \gG_n^{(t)})}{P(X_S \in \gG_n^{(t)})} \bigg| \E_{p_T}\left[\frac{\textbf{1}(Y(X) = \hat{Y}(X)) - \hat{\alpha}_T(\gG_n^{(t)}; w^\dagger(n))}{w^\dagger(n)(X)} \big| \gG_n^{(t)} \right] \bigg| 
\end{align}
\begin{align}
    & \leq \frac{P(X_T \in \gG_n^{(t)})}{P(X_S \in \gG_n^{(t)})} \left( \E_{p_T}\left[(\textbf{1}(Y(X) = \hat{Y}(X)) - \hat{\alpha}_T(\gG_n^{(t)}; w^\dagger(n)))^2 \big| \gG_n^{(t)} \right] \E_{p_T}\left[\frac{1}{w^\dagger(n)(X)^2} \big| \gG_n^{(t)} \right] \right)^{1/2} 
    \label{eq:sep_by_cauchy} \\  
    & \leq \frac{P(X_T \in \gG_n^{(t)})}{2P(X_S \in \gG_n^{(t)})} \left( \E_{p_T}\left[(\textbf{1}(Y(X) = \hat{Y}(X)) - \hat{\alpha}_T(\gG_n^{(t)}; w^\dagger(n)))^2 \big| \gG_n^{(t)} \right] + \E_{p_T}\left[\frac{1}{w^\dagger(n)(X)^2} \big| \gG_n^{(t)} \right] \right)
    \label{eq:sep_by_amgm}  \\
    & \leq \frac{P(X_T \in \gG_n^{(t)})}{2P(X_S \in \gG_n^{(t)})} \left( \E_{p_T}\left[(\textbf{1}(Y = \hat{Y}) - \hat{\alpha}_T(\gG_n^{(t)}; w^\dagger(n)))^2 \big| \gG_n^{(t)} \right] + \frac{1}{\ubar{w}^\dagger(n)^2} \right) 
\end{align}
where \Eqref{eq:sep_by_cauchy} holds due to the Cauchy-Schwarz inequality, \Eqref{eq:sep_by_amgm} holds due to the AM-GM inequality, and $\ubar{w}^\dagger(n) := \min_{i \in [2B]} \{ w^\dagger_i(n) \}$.
\end{proof}

\subsection{Formal Statement and Proof of Proposition \ref{proposition:bias_var_decomp}}
\begin{proposition}[Bias-variance decomposition] \label{proposition:bias_var_decomp}
    Let $\hat{\alpha}_T(\gG_n^{(t)})$ be an estimate for $\alpha_T(\gG_n^{(t)}; w^*)$. Then, the bias of the identical accuracy assumption is given by 
    \begin{equation} 
        IdentBias(w^\dagger(n); \gG_n^{(t)}) 
        =\tfrac{P(X_T \in \gG_n^{(t)})}{2P(X_S \in \gG_n^{(t)})}  \left( \tfrac{1}{\ubar{w}^\dagger(n)^2} + Bias(\hat{\alpha}_T(\gG_n^{(t)}))^2 + Var(\textbf{1}(Y = \hat{Y}) | \gG_n^{(t)} ) \right)
    \end{equation}
    where $Bias(\hat{\alpha}_T(\gG_n^{(t)})) := | \alpha_T(\gG_n^{(t)}; w^*) - \hat{\alpha}_T(\gG_n^{(t)}) |$ is the bias of the estimate $\hat{\alpha}_T(\gG_n^{(t)})$ and $ Var(\textbf{1}(Y = \hat{Y}) | \gG_n^{(t)} ) := \E_{p_T}\left[ \left( \textbf{1}(Y = \hat{Y}) - \alpha_T(\gG_n^{(t)}; w^*) \right)^2 | \gG_n^{(t)} \right] $ is the variance of the correctness of predictions in $\gG_n^{(t)}$.
\end{proposition}
\begin{proof}    
    Based on the proof of Proposition~\ref{proposition:closeness_opt}, it is enough to decompose $\E_{p_T} [(\textbf{1}(Y(X) = \hat{Y}(X)) - \hat{\alpha}_T(\gG_n^{(t)})  ]^2$ as follows
    \begin{align}
        &  \E_{p_T} \left[ ( \textbf{1}(Y(X) = \hat{Y}(X)) - \hat{\alpha}_T(\gG_n^{(t)})  )^2 | \gG_n^{(t)} \right] \\
        & = \E_{p_T} \left[  ( \textbf{1}(Y(X) = \hat{Y}(X)) - \alpha_T(\gG_n^{(t)}; w^*) + \alpha_T(\gG_n^{(t)}; w^*) - \hat{\alpha}_T(\gG_n^{(t)})  )^2 | \gG_n^{(t)} \right] \\ 
        & = \E_{p_T} \left[ ( \textbf{1}(Y(X) = \hat{Y}(X)) - \alpha_T(\gG_n^{(t)}; w^*) )^2 + ( \alpha_T(\gG_n^{(t)}; w^*) - \hat{\alpha}_T(\gG_n^{(t)})  )^2 | \gG_n^{(t)} \right] \label{eq:cross_term} 
    \end{align}
    where the equality \eqref{eq:cross_term} holds due to $\E_{p_{T_X}} \E_{p_{T_{Y|X}}}[ (\textbf{1}(Y(X) = \hat{Y}(X)) - \alpha_T(\gG_n^{(t)}; w^*))(\alpha_T(\gG_n^{(t)}; w^*) - \hat{\alpha}_T(\gG_n^{(t)})) | \gG_n^{(t)} ] = 0 $.
\end{proof}

\section{Discussions}

\subsection{Model Calibration in the I.I.D. Settings} \label{appx:literature_calib_iid}
In a classification problem, the maximum value of the softmax output is often considered as a confidence of a neural network's prediction. In \citep{guo2017calibration}, it is shown that the modern neural networks are poorly calibrated, tending to produce larger confidences than their accuracies. Based on this observation, \citep{guo2017calibration} introduce a post-processing approach that adjusts a temperature parameter of the softmax function for adjusting the overall confidence level. In Bayesian approaches (such as Monte-Carlo dropout \citep{gal2016dropout, gal2017concrete}, deep ensemble \citep{lakshminarayanan2017simple, rahaman2021uncertainty}, and a last-layer Bayesian approach \citep{sensoy2018evidential,joo2020being}), the confidence level adjustment is induced by posterior inference and model averaging. While both post-hoc calibration methods and Bayesian methods have been achieving impressive calibration performances in the i.i.d. setting \citep{maddox2019simple, ovadia2019can, ebrahimi2019uncertainty}, it has been shown that most of the calibration improvement methods fall short under distribution shifts \citep{ovadia2019can}.

\subsection{On Choice of Non-Parametric Estimators} \label{appx:discussion_non_params}
Our concept of determining the IW from its CI can be applied to any other valid CI estimators. For example, by analyzing a CI of the odds ratio of the logistic regression used as a domain classifier \citep{bickel2007discriminative,park2020calibrated,salvador2021frustratingly}, a CI of the IW can be obtained. Then, IW-GAE can be applied in the same way as developed in Section~\ref{sec:new_method}. As an extreme example, we apply IW-GAE by setting minimum and maximum values of IWs as CIs in an ablation study (Table \ref{table:ablation}). While IW-GAE outperforms strong baseline methods (CPCS and TransCal) even under this naive CI estimation, we observe that its performance is reduced compared to the setting with a sophisticated CI estimation \citep{park2021pac}. In this regard, advancements in IW estimation or CI estimation would be beneficial for accurately estimating the group accuracy, thereby model selection and uncertainty estimation. Therefore, we leave combining IW-GAE with advanced IW estimation techniques as an important future direction of research. 

\subsection{On Choice of the Number of Groups} \label{appx:discussion_num_groups}
In this work, we estimate the group accuracy by grouping predictions based on the confidence of the prediction. Therefore, a natural question to ask is how to select the number of groups. If we use a small number of groups, then there would be high $IdentBias(w^\dagger; \gG_n^{(t)})$ because of the large variance of prediction correctness within a group. In addition, reporting the same accuracy estimate for a large number of predictions could be inaccurate in terms of representing uncertainty for individual predictions. Conversely, if we use a large number of bins, there would be high Monte-Carlo approximation errors, $\epsilon_{stat}$. Therefore, it would result in a loose connection between the source group accuracy estimation error and the objective in the optimization problem (cf. Proposition~\ref{proposition:closeness_opt}). Based on our experimental results with $M=10$ and the sensitivity analysis (cf. Section \ref{sec:senst_analy}), we would recommend to choose $M$ from its standard range $[10,20]$ in the literature.

\section{Additional Details}

\subsection{Obtaining CIs of Binned IWs} \label{appx:cp_ci_explain}
In this work, we use a recently proposed nonparametric estimation method \citep{park2021pac} for constructing the CI of the IW\footnote{We note that IW-GAE works with any interval estimation methods (cf. Discussions in Appendix \ref{appx:discussion_non_params}) or even arbitrary intervals (cf. Section \ref{sec:ablation}).}. In this approach, $\gX$ is partitioned into $B$ number of bins ($\gX = \cup_{i=1}^{B} \gB_i$) where $\gB_i = \{x \in \gX | I^{(B)}(x) = i \}$. Here, the partition function is defined such that $I^{(B)}(x) = j$ if $q(\frac{j-1}{B}; \gW) < \tilde{w}(x) \leq q(\frac{j}{B}; \gW)$ where $\gW := \{\tilde{w}(x) | x  \in \gD_S \cup \gD_T \}$ and $\tilde{w}$ is a rough estimate of IW (cf. Appendix \ref{appx:iw_estim_logistic}).

Then, confidence intervals (CIs) of the binned probabilities $\bar{p}_S(x) = \bar{p}_{S_{I^{(B)}(x)}} \text{ with } \bar{p}_{S_j} = \int_{\gB_j} p_S(x) dx $ and $\bar{p}_T(x) = \bar{p}_{T_{I^{(B)}(x)}} \text{ with } \bar{p}_{T_j} = \int_{\gB_j} p_T(x) dx$ are constructed. Specifically, for $\bar{p}_{S_j}$, the number of samples in a bin $n^{(S)}_j := \sum_{i=1}^{N^{(S)}} \textbf{1}(x^{(S)}_i \in \gB_j)$ is interpreted as a sample from $\text{Binom}(N^{(S)}, \bar{p}_{S_j})$. Then, the Clopper–Pearson CI \citep{clopper1934use} provides the CI of $\bar{p}_{S_j}$ as $\ubar{\theta}_S (n^{(S)}_j; N^{(S)}, \delta/2) \leq \bar{p}_{S_j} \leq \bar{\theta}_S (n^{(S)}_j; N^{(S)}, \delta/2)$ with probability at least $1-\delta$ where $\bar{\theta}(k; m, \delta) := \inf \{\theta \in [0,1] | F(k; m, \theta) \leq \delta \} $ and $\ubar{\theta}(k; m, \delta) := \sup \{\theta \in [0,1] | F(k; m, \theta) \geq \delta \} $ with $F$ being the cumulative distribution function of the binomial distribution. Similarly, we can obtain the CIs of $\bar{p}_{T_j}$ by collecting $n^{(T)}_j := \sum_{i \in [N^{(T)}]} \textbf{1}(x_i^{(T)} \in \gB_j)$ and following the same procedure, which are denoted as $\ubar{\theta}_T (n^{(T)}_j; N^{(T)}, \delta/2) $ and $\bar{\theta}_T (n^{(T)}_j; N^{(T)}, \delta/2) $.

With the CIs of $p_{S_j}$ and $p_{T_j}$ for $j \in [B]$, the CI of the IW in $\gB_j$ can be obtained. Specifically, for $ \bar{\delta} := \delta / 2B$, the following inequality holds with probability at least $1 - \delta$ \citep{park2021pac}:
\begin{equation} \label{eq:ci_imp_weight}
    \frac{[\ubar{\theta}_T (n^{(T)}_j; N^{(T)}, \bar{\delta}) - G]^+}{\bar{\theta}_S (n_j^{(S)}; N^{(S)}, \bar{\delta}) + G} \leq w_j^* := \frac{\bar{p}_{T_j}}{\bar{p}_{S_j}} \leq \frac{\bar{\theta}_T (n^{(T)}_j; N^{(T)}, \bar{\delta}) + G}{[\ubar{\theta}_S (n_j^{(S)}; N^{(S)}, \bar{\delta}) - G]^+}
\end{equation}
where $[a]^+ := \max\{0, a\}$ and  $G \in \R_+$  is a constant that satisfies $\int_{\gB_j} |p_S(x) - p_S(x^\prime)| dx^\prime \leq G $ and $\int_{\gB_j} |p_T(x) - p_T(x^\prime)| dx^\prime \leq G $ for all $x \in \gB_j$ and $j \in [B]$. We refer to $\{w_i^* \}_{i \in B}$ as \textbf{binned IWs}. Also, we define the CI of $w^*_j$ as
\begin{equation}
    \Phi_j := \left[\frac{[\ubar{\theta}_T (n^{(T)}_j; N^{(T)}, \bar{\delta}) - G]^+}{\bar{\theta}_S (n_j^{(S)}; N^{(S)}, \bar{\delta}) + G}, \frac{\bar{\theta}_T (n^{(T)}_j; N^{(T)}, \bar{\delta}) + G}{[\ubar{\theta}_S (n_j^{(S)}; N^{(S)}, \bar{\delta}) - G]^+} \right].
\end{equation}

\subsection{IW Estimator}  \label{appx:iw_estim_logistic}
IW estimation is required for implementing baseline methods and construct bins for estimating the CI of the IW. We adopt a logistic regression model on top of the neural network's representation as the discriminative learning-based estimation \citep{bickel2007discriminative}, following \citet{wang2020transferable}. Specifically, it first upsamples from one domain to make $|\gD_S| = |\gD_T|$, and then it labels samples with the domain index: $\{(h(x),1) | x \in \gD_T \}$ and $\{(h(x),0) | x \in \gD_S \}$ where $h$ is the feature map of the neural network. After training the logistic regression model $v$ with a quasi-Newton method until convergence, the IW can be estimated as $\tilde{w}(x) = v(h(x)) / (1-v(h(x)))$.

\subsection{Algorithm} \label{appx:gae_algorithm}

\begin{algorithm}[H]
\caption{Pseudocode of IW-GAE \label{alg:iw_gae}}
\begin{algorithmic}
\STATE \textbf{Input: } Source dataset $\gD_S = \{(x_i^{(S)}, y_i^{(S)})\}_{i=1}^{N^{(S)}}$, Target dataset $\gD_T = \{x_i^{(T)}\}_{i \in [N^{(T)}]}$
\STATE \textbf{Hyperparameters: } The numbers of bins and groups ($B$ and $M$), level of CI $\bar{\delta}$, search space $\gT$
\STATE \textit{\# Prepare a UDA model \citep{wang2020transferable}}
\STATE Partition $\gD_S$ into $\gD_S^{tr}$ and $\gD_S^{val}$
\STATE Train a neural network $g$ on $(\gD_S^{tr}, \gD_{T})$ with any UDA method 
\STATE Upsample $\gD_S^{tr}$ or $\gD_T$ to make $|\gD_S^{tr}| = | \gD_T |$ 
\STATE Compute  $\gF_S^{tr} = \{g(x) | x \in \gD_S^{tr}\}$, $\gF_S^{val} = \{g(x) | x \in \gD_S^{val}\}$, and $\gF_T = \{g(x) | x \in \gD_T\}$
\item Train a logistic regression model $H$ that discriminates $\gF_S^{tr}$ and $\gF_T$ 
\STATE \textit{\# Obtain CIs of binned IWs \citep{park2021pac} }
\STATE Gather IWs $\gW_{S \cup T} = \{(1 - H(g(x))) / H(g(x)): x \in \gD_S^{val} \cup \gD_T \}$
\STATE Compute quantiles $q(i) = i/(B+1)$-th quantile of $\gW_{S \cup T}$ for $ i \in [B+1]$
\STATE Construct bins $\gB_i = \{x \in  \gD_S^{val} \cup \gD_T: q(i) \leq (1 - H(g(x))) / H(g(x)) \leq q(i+1) \}$ for $i \in [B]$
\STATE Compute $\Phi_i$ using \eqref{eq:ci_imp_weight} for each $i \in [B]$
\STATE \textit{\# IW-GAE }
\STATE $f^\dagger = \infty$ 
\FOR{$t \in \gT$}
\STATE Obtain $w(n; t)$ by solving the optimization problem in  \eqref{eq:opt_prob}-\eqref{eq:const2} for $n \in [M]$ 
\IF{$\sum_{n \in [M]}  \left( \hat{\alpha}_S^{(MC)}(\gG_n^{(t)}) - \hat{\alpha}_S^{(IW)}(\gG_n^{(t)}; w(n; t)) \right)^2 \leq f^\dagger$}
\STATE $f^\dagger = \sum_{n \in [M]}  \left( \hat{\alpha}_S^{(MC)}(\gG_n^{(t)}) - \hat{\alpha}_S^{(IW)}(\gG_n^{(t)}; w(n; t)) \right)^2$
\STATE $t^\dagger = t$
\STATE $w^\dagger(n; t^\dagger) = w(n; t)$ for $n \in [M]$
\ENDIF
\ENDFOR
\end{algorithmic}
\end{algorithm}

\section{Experimental Details} \label{appx:add_exp_details}
We follow the exact same training configurations as those used in the Transfer Learning Library \citep{tllib}, except we separate 20\% as the validation dataset from the source domain (in the original implementation, validation is performed with the test dataset for OfficeHome). The configuration of training MDD for OfficeHome is as follows: MDD is trained for 30 epochs with SGD with momentum parameter 0.9 and weight decay of $0.0005$. The learning rate is schedule by $\alpha \cdot (1 + \gamma \cdot t)^{- \eta}$ where $t$ is the iteration counter, $\alpha = 0.004$, $\gamma = 0.0002$, $\eta = 0.75$, and the stochastic gradient is computed with minibatch of 32 samples from the source domain and 32 samples from the target domain. Also, it uses the margin coefficient of 4 as the MDD-specific hyperparamter. For the model architecture, it uses ResNet-50 pre-trained on ImageNet \citep{russakovsky2015imagenet} with the bottleneck dimension of 2,048. For the large-scale VisDa-2017, only the architecture is changed to ResNet-101 with the bottleneck dimension of 1,024 under the same training configuration.

\paragraph{IW-GAE specific details}
For CI estimation, we follow the same configuration with the original method \citep{park2021pac}. Specifically, we use constant $G = 0.001$, CI level $\bar{\delta} = 0.05$, and the number of bins $B=10$. In addition, we use the maximum IW value $\tilde{w}_n^{(ub)} = 6.0$ and the minimum IW value $\ubar{w}^\dagger(n) = 1/6$ for $n \in [M]$, which is a common technique in IW-based estimations \citep{wang2020transferable,park2021pac}. For the constraint relaxation constants in IW-GAE, we use $\delta_{tol} = 0.1$ and $\delta_{pr} = 0.3$. For implementing sequential least square programming, we use the SciPy Library \citep{2020SciPy} with tolerance $10^{-8}$ that is used to check a convergence condition (other optimizer-specific values follow the default values in SciPy) and choose the middle points from CIs of binned IW as an initial solution.

\newpage 

\section{Missing Tables} \label{appx:missing_tables}

\begin{table}[h!]
\vskip -0.1in
\caption{Model calibration benchmark results of MDD with ResNet-50 on Office-Home. We repeat experiments for ten times and report the average value of ECE. }
\vskip 0.1in
\scalebox{0.85}{
\begin{tabular}{l|rrrrrrrrrrrr|r}
\hline
Method  & Ar-Cl & Ar-Pr & Ar-Rw & Cl-Ar & Cl-Pr & Cl-Rw & Pr-Ar & Pr-Cl & Pr-Rw & Rw-Ar & Rw-Cl & Rw-Pr & Avg \\ \hline
Vanilla  & 40.61 &        25.62 &         15.56 &         33.83 &         25.34 &         24.75 &         33.45 &         38.62 &         16.76 &         23.37 &         36.51 &         14.01 &         27.37 \\
TS  & 35.86 &        22.84 &         10.60 &          28.24 &         20.74 &         20.06 &         32.47 &         37.20 &          14.89 &         18.36 &         34.62 &         12.28 &         24.01 \\
CPCS      & 22.93 &        22.07 &         10.19 &         26.88 &         18.36 &         14.05 &         28.28 &         29.20 &          12.06 &         15.76 &         26.54 &         11.14 &         19.79 \\
IW-TS   & 32.63 &        22.90 &          11.27 &         28.05 &         19.65 &         18.67 &         30.77 &         38.46 &         15.10 &          17.69 &         32.20 &          11.77 &         23.26 \\
TransCal   & 33.57 &        20.27 &          {8.88} &          26.36 &         18.81 &         18.42 &         27.35 &         29.86 &         10.48 &         16.17 &         29.90 &          10.00 &          20.84 \\
PTS & 31.91 & 24.36 & 10.65 & 22.81 & 20.42 & 15.92 & 26.55 & 39.34 & 12.38 & 21.83 & 26.31 & 10.44 & 21.91 \\
AvUTS & 29.59 & 25.55 & 10.40 & 31.81 & 26.06 & 26.15 & 37.10 & 46.04 & 20.75 & 27.70 & 41.27 & 15.61 & 28.17 \\
IW-Mid  & 23.25 &        31.62 &         12.99 &         17.15 &         18.71 &         9.23 &          27.75 &         30.35 &         9.02 &          13.64 &         26.32 &         10.60 &          19.22 \\
IW-GAE  &  {12.78} &         {4.70} &   12.93 &          {7.52} &           {4.42} &           {4.11} &          { 9.50} &    {17.49 }&          {8.40} &    {7.62} &           {9.52} &           {8.14} &           {8.93} \\ \hline
Oracle & 10.45 &        10.72 &         6.47 &          8.10 &   7.62 &          6.55 &          11.88 &         9.39 &          5.93 &          7.54 &          10.72 &         5.70 &   8.42 \\ \hline
\end{tabular}}
\label{table:full_office_home_ece}
\vskip -0.1in
\end{table}

\begin{table}[h!]
\vskip -0.1in
\caption{Large-scale model calibration benchmark results of CDAN with ResNet-50 on DomainNet. The numbers indicate the mean ECE across ten repetitions. Cl, Pt, Rw, and Sk correspond to clipart, painting, real, and sketch, respectively. }
\vskip 0.1in
\scalebox{0.85}{
\begin{tabular}{l|rrrrrrrrrrrr|r}
\hline
Method  & Cl-Pt &    Cl-Rw &   Cl-Sk &   Pt-Cl &   Pt-Rw &   Pt-Sk &   Rw-Cl &   Rw-Pt &   Rw-Sk &   Sk-Cl &   Sk-Pt &   Sk-Rw & Avg \\ \hline
Vanilla & 13.23 &        6.36 &          12.92 &         9.75 &          6.35 &          15.56 &         9.44 &          9.70 &   14.34 &         6.63 &          11.25 &          {5.23} &          10.06 \\
TS & 12.95 &         {5.95} &          13.32 &         6.40 &   3.90 &          11.07 &         8.64 &          10.49 &         16.08 &          {3.17} &          5.58 &          13.09 &         9.22 \\
CPCS &  {5.64} &         21.90 &          7.70 &    {5.14} &          7.72 &          7.90 &   9.35 &          11.17 &         17.06 &         3.46 &           {2.23} &          15.90 &          9.60 \\
IW-TS & 16.76 &        16.70 &          12.53 &         5.29 &          7.84 &          4.34 &          9.60 &   10.58 &         16.80 &          5.40 &   2.98 &          17.11 &         10.49 \\
TransCal & 18.51 &        29.63 &         20.92 &         23.02 &         31.83 &         17.58 &         27.88 &         28.83 &         20.31 &         31.66 &         23.06 &         31.46 &         25.39 \\
IW-Mid & 7.61 &         11.01 &         5.89 &          8.84 &          7.58 &          5.36 &          8.70 &   7.49 &          7.53 &          10.24 &         8.10 &   10.21 &         8.21 \\
IW-GAE & 6.06 &         8.15 &           {5.38} &          7.45 &           {3.89} &           {3.94} &           {7.01} &           {5.58} &           {6.73} &          6.80 &   6.82 &          8.00 &    {6.32} \\ \cline{1-14}
Oracle & 4.55 &         2.78 &          4.01 &          3.10 &   3.72 &          2.72 &          3.10 &   2.79 &          2.83 &          3.13 &          1.70 &   1.77 &          3.02 \\ \hline
\end{tabular}}
\label{table:full_domain_net}
\vskip -0.1in
\end{table}

\begin{table}[h!]
\vskip -0.1in
\caption{Hyperparameter selection benchmark results of MDD with ResNet-50 on Office-Home. We repeat experiments for ten times and report the average test accuracy of selected model.}
\vskip 0.1in
\scalebox{0.82}{
\begin{tabular}{l|rrrrrrrrrrrr|r}
\hline
Method  & Ar-Cl & Ar-Pr & Ar-Rw & Cl-Ar & Cl-Pr & Cl-Rw & Pr-Ar & Pr-Cl & Pr-Rw & Rw-Ar & Rw-Cl & Rw-Pr & Avg \\ \hline
Vanilla  & 53.31 &         {70.96} &         77.44 &         59.70 &          65.17 &         69.96 &         57.07 &         50.95 &         74.75 &         68.81 &         57.11 &         80.13 &         65.45 \\
IWCV  & 53.24 &        69.61 &         72.50 &          59.70 &          65.17 &         67.50 &          57.07 &         55.21 &         74.75 &         68.81 &         58.51 &         80.13 &         65.18 \\
DEV   & 53.31 &        70.72 &         77.44 &         59.79 &         67.99 &         69.96 &         57.07 &         52.50 &          77.12 &          {70.50} &          53.38 &         82.27 &         66.00 \\
InfoMax           &  {54.34}     &  {70.96}      & 77.53      &  {61.48}      &  {69.93}      &  {71.06}      &  {62.79}      & 54.41      &  {78.79}      &  {71.32}      & 58.51      & 82.36      & 67.79      \\
SND          & 44.55      & 68.14      & 75.57      & 58.86      & 66.04      & 66.46      & 61.13      & 53.20      & 70.38      & 62.54      & 56.15      & 80.20      & 63.60      \\
Corr-C       & 50.88      &  {70.96}      &  {78.47}      & 60.74      & 68.60      &  {71.06}      &  {62.79}      & 54.41      &  {78.79}      &  {71.32}      &  {58.51}      & 82.36      & 67.41      \\
TransScore   &  {54.34}      &  {70.96}      & 77.53      &  {61.48}      &  {69.93}      &  {71.06}      &  {62.79}      & 54.41      &  {78.79}      &  {71.32}      &  {58.51}      & 82.36      & 67.87      \\
MixVal       &  {54.34}      & 70.09      & 77.43      & 60.74      & 59.34      & 70.17      & 61.00      &  {55.21}      & 78.35      &  {71.32}      & 57.89      & 82.36      & 66.52      \\
IW-Mid  & 54.13 &        69.27 &          {78.47} &          {61.48} &         68.03 &          {71.06} &         59.99 &         55.21 &         78.79 &          {70.50} &          57.11 &         83.10 &          67.26 \\
IW-GAE  &  {54.34} &         {70.96} &          {78.47} &          {61.48} &          {69.93} &          {71.06} &          {62.79} &          {55.21} &          {78.79} &          {70.50} &           {58.51} &          {83.31} &          {67.95} \\ \hline
Lower bound &  52.51 &        69.27 &         72.50 &          59.70 &          65.17 &         67.50 &          57.07 &         50.95 &         74.75 &         68.81 &   50.90 &          80.13 &         64.10 \\
\hline 
Oracle & 54.34 &        70.96 &         78.47 &         61.48 &         69.93 &         71.06 &         62.79 &         55.21 &         78.79 &         71.32 &         58.51 &         83.31 &         68.01 \\ \hline
\end{tabular}}
\label{table:full_hyperparam}
\vskip -0.1in
\end{table}

\begin{table}[h!]
\vskip -0.1in
\caption{Checkpoint selection benchmark results of MDD with ResNet-50 on Office-Home. We repeat experiments for ten times and report the average test accuracy of selected model.  }
\vskip 0.1in
\scalebox{0.82}{
\begin{tabular}{l|llllllllllll|l}
\hline
Method  & Ar-Cl & Ar-Pr & Ar-Rw & Cl-Ar & Cl-Pr & Cl-Rw & Pr-Ar & Pr-Cl & Pr-Rw & Rw-Ar & Rw-Cl & Rw-Pr & Avg \\ \hline
Vanilla  & 47.22 &        74.14 &         77.76 &         61.85 &         70.96 &         71.59 &         60.98 &  53.63 &        {78.93} &         71.57 &         57.04 &         83.96 &         67.47 \\

IWCV &  {54.46} &         {74.22} &         72.27 &         61.48 &         70.49 &         70.62 &         61.30 &         51.13 &         78.37 &         72.94 &         58.43 &         84.00 &         67.48 \\
DEV & 54.04 &        73.94 &         78.16 &         61.52 &         63.19 &         70.70 &          60.43 &         53.63 &         78.93 &         71.57 &         58.62 &         83.89 &         67.39 \\
InfoMax & 54.32 & 74.72 & 77.90 &  {62.79} & 71.03 & 71.47 &  {61.39} & 53.15 & 78.75 & 72.89 & 58.53 & 83.94 & 68.38 \\
SND & 47.67 & 73.06 & 77.71 & 62.67 & 70.47 & 71.17 & 61.43 & 52.14 & 78.75 & 70.71 & 57.66 & 79.80 & 67.23 \\
Corr-C & 54.46 &  {74.72} & 77.53 & 61.76 & 70.88 & 71.24 & 61.30 & 52.47 & 78.40 & 72.59 & 58.53 & 83.89 & 68.24 \\
TransScore &  {54.79} & 74.14 & 77.77 & 61.76 & 70.97 & 71.48 & 61.17 & 53.15 &  {78.93} & 72.89 & 58.53 & 83.89 & 68.38 \\
MixVal & 54.45 & 74.35 & 77.77 & 61.93 & 70.70 & 71.43 & 61.30 & 53.29 &  {78.93} & 72.89 & 58.53 & 83.89 & 68.37 \\
IW-Mid  & 54.04 &        72.63 &         78.37 &          {62.05} &          {71.28} &         71.45 &          {61.25} &          {54.39} &          {79.07} &         73.19 &          {58.75} &         80.06 &         68.04 \\

IW-GAE & 54.32 &        73.98 &          {78.51} &         61.96 &         71.25 &          {71.70} &          61.10 &          54.30 &          78.91 &          {73.22} &         58.70 &           {83.86} &          {68.48} \\ \hline
Lower bound  & 41.90 &         64.88 &         72.27 &         52.00 &          58.48 &         62.13 &         53.52 &       38.33 &  70.92 &         63.41 &         44.81 &         75.83 &         58.21 \\   \hline
Oracle & 54.80 &         74.79 &         78.61 &         62.79 &         71.59 &         72.18 &         61.64 &      54.64 &   79.44 &         73.42 &         59.43 &         84.12 &         68.95 \\ \hline
\end{tabular}}
\label{table:full_check_pt}
\vskip -0.1in
\end{table}

\newpage 

\section{Additional Experiments} \label{appx:cdan_exp}
In this section, we show the effectiveness of IW-GAE with two different base models. First, we perform additional experiments with conditional domain adversarial network (CDAN; \citep{long2018conditional}) which is also a popular UDA method. As in the experiments with MDD, we use ResNet-50 as the backbone network and OfficeHome as the dataset. The learning rate schedule for CDAN is $\alpha \cdot (1 + \gamma \cdot t)^{- \eta}$ where $t$ is the iteration counter, $\alpha = 0.01$, $\gamma = 0.001$, and $\eta = 0.75$. The remaining training configuration for CDAN is the same as the MDD training configuration except it uses the bottleneck dimension of 256 and weight decay of $0.0005$ (cf. Appendix \ref{appx:add_exp_details}). As we can see from Table \ref{table:add_cdan_ece}, IW-GAE achieves the best performance among all considered methods, achieving the best ECE in 8 out of the 12 cases as well as the lowest mean ECE. We note that TransCal achieves a performance comparable to IW-GAE in this experiment, but considering the results in the other tasks, IW-GAE is still an appealing method for performing the model calibration task.

We also perform additional experiments with maximum classifier discrepancy (MCD; \citep{saito2018maximum}). Following the previous experiments with MDD and CDAN, we use ResNet-50 as the backbone network and OfficeHome as the dataset. The training configuration is the same as the MDD training configuration except it uses the fixed learning rate of 0.001 with weight decay of $0.0005$ and bottleneck dimension of 1,024 (cf. Appendix \ref{appx:add_exp_details}). Consistent to other benchmark results, IW-GAE achieves the best performance among all methods (Table \ref{table:add_mcd_ece}). Specifically, IW-GAE achieves the best average model calibration performance, and its ECE is lowest in 7 out of 12 domain pairs. Note that IW-Mid's performance with MCD is significantly lower compared to other benchmark results. However, IW-GAE still significantly improves the performance, indicating that IW-GAE does not strongly depends on accuracy of the CI estimation discussed in Section \ref{appx:cp_ci_explain}.

\begin{table}[]
\vskip -0.1in
\caption{Model calibration benchmark results of CDAN  with ResNet-50 on Office-Home. The numbers indicate the mean ECE across ten repetitions. }
\vskip 0.1in
\scalebox{0.85}{
\begin{tabular}{l|rrrrrrrrrrrr|r}
\hline
Method  & Ar-Cl & Ar-Pr & Ar-Rw & Cl-Ar & Cl-Pr & Cl-Rw & Pr-Ar & Pr-Cl & Pr-Rw & Rw-Ar & Rw-Cl & Rw-Pr & Avg \\ \hline
Vanilla  & 30.73 &        18.38 &         14.37 &         25.63 &         22.44 &         19.10 &          27.54 &         36.72 &         12.48 &     19.93 &         31.12 &         10.88 &         22.44 \\
TS  & 29.68 &        19.40 &          14.40 &          22.15 &         19.97 &         16.88 &         28.82 &         38.03 &         12.99 &     20.46 &         31.91 &         11.83 &         22.21 \\
CPCS      & 18.78 &        18.09 &         14.74 &         22.18 &         20.74 &         16.33 &         29.30 &          34.92 &         11.92 &         20.99 &         31.41 &         11.07 &         20.87 \\
IW-TS   & 12.38 &        16.79 &         14.85 &         21.75 &         20.06 &         16.92 &         29.30 &          38.84 &         13.30 &          20.82 &         31.10 &          11.37 &         20.62 \\
TransCal   &  {7.94} &          {14.05} &         12.91 &         7.82 &          9.25 &          10.23 &         9.37 &          12.60 &          14.29 &          {9.92} &          9.76 &          17.51 &         11.30 \\
IW-Mid  & 36.05 &        47.70 &          26.82 &         21.08 &         22.95 &         21.55 &         18.88 &         28.99 &         15.39 &         21.16 &         28.16 &         25.27 &         26.17 \\
IW-GAE & 13.98 &        29.82 &          {9.44} &           {6.55} &           {5.59} &           {10.16} &          {5.29 }&          13.47 &          {11.01} &         11.12 &          {7.26} &           {9.84 }&           {11.13 }\\ \cline{1-14}
Oracle  & 7.91 &         8.80 &   6.05 &          7.57 &          7.93 &          6.76 &          9.07 &          9.14 &          4.04 &          7.16 &          9.19 &          5.65 &          7.44  \\\hline
\end{tabular}}
\label{table:add_cdan_ece}
\vskip -0.1in
\end{table}

\begin{table}[]
\vskip -0.1in
\caption{Model calibration benchmark results of MCD with ResNet-50 on Office-Home. The numbers indicate the mean ECE across ten repetitions. }
\vskip 0.1in
\scalebox{0.85}{
\begin{tabular}{l|rrrrrrrrrrrr|r}
\hline
Method  & Ar-Cl & Ar-Pr & Ar-Rw & Cl-Ar & Cl-Pr & Cl-Rw & Pr-Ar & Pr-Cl & Pr-Rw & Rw-Ar & Rw-Cl & Rw-Pr & Avg \\ \hline
Vanilla & 38.91 &        26.39 &         18.86 &         32.85 &         26.69 &         19.36 &         35.87 &         36.70 &         18.61 &         24.57 &         36.87 &         14.79 &         27.54 \\
TS & 31.84 &        22.55 &         13.49 &         26.16 &         20.10 &         10.72 &         33.98 &         31.91 &         15.59 &         21.62 &         31.59 &         12.46 &         22.67 \\
CPCS & 13.07 &        20.09 &         47.15 &          {9.78} &          21.82 &         8.02 &          32.65 &         25.61 &         15.27 &         20.53 &         40.38 &         7.84 &          21.85 \\
IW-TS &  {12.88} &        21.44 &         61.15 &         10.56 &         16.40 &         11.72 &         33.03 &         36.37 &         14.09 &         19.96 &         41.95 &         19.30 &         24.91 \\
TransCal & 19.23 &        15.09 &         6.55 &          17.91 &         11.60 &         3.91 &           {22.98} &          {15.81} &          {6.11} &          13.77 &         21.40 &         4.02 &          13.20 \\
IW-Mid & 50.68 &        28.93 &         23.92 &         38.24 &         33.48 &         28.58 &         39.76 &         37.45 &         22.40 &         27.15 &         44.15 &         18.07 &         32.73 \\
IW-GAE & 22.21 &         {10.68} &          {2.38} &          15.96 &          {9.30} &           {3.53} &          23.54 &         22.73 &         6.37 &           {11.78} &          {20.75} &          {1.63} &           {12.57} \\ \cline{1-14}
Oracle & 5.88 &         9.91 &          3.19 &          7.75 &          4.64 &          3.66 &          4.17 &          7.70 &          3.09 &          4.51 &          8.09 &          3.54 &          5.51 \\ \hline
\end{tabular}}
\label{table:add_mcd_ece}
\vskip -0.1in
\end{table}

\subsection{Qualitative Evaluation of IW-GAE} \label{appx:qualit_eval}
\begin{figure}
% \vskip -0.1in
    \centering{
    \begin{subfigure}
     {\includegraphics[width=0.23\textwidth]{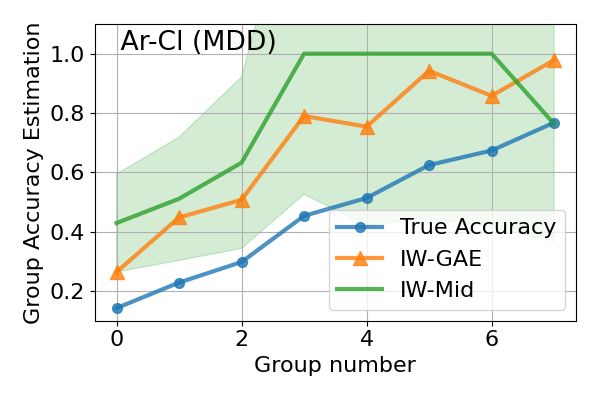}}
    \end{subfigure}
    \begin{subfigure}
     {\includegraphics[width=0.23\textwidth]{experiments/MDD_Ar-Pr.png}}
    \end{subfigure}
    \begin{subfigure}
     {\includegraphics[width=0.23\textwidth]{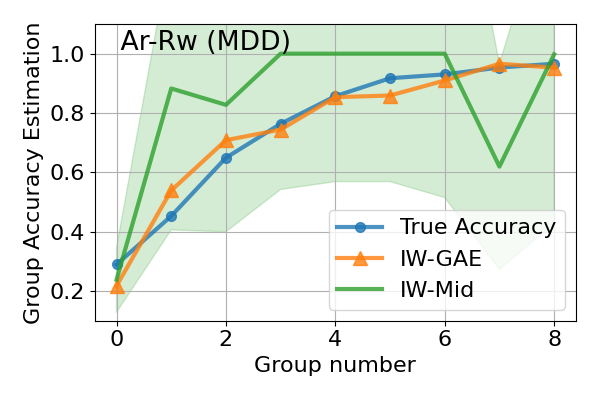}}
    \end{subfigure}
    \begin{subfigure}
     {\includegraphics[width=0.23\textwidth]{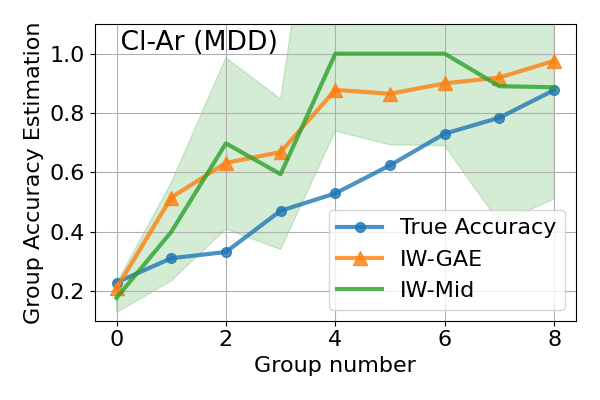}}
    \end{subfigure} \\
    \begin{subfigure}
     {\includegraphics[width=0.23\textwidth]{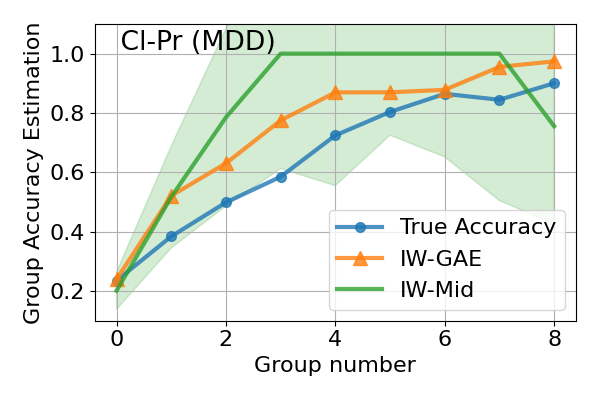}}
    \end{subfigure}
    \begin{subfigure}
     {\includegraphics[width=0.23\textwidth]{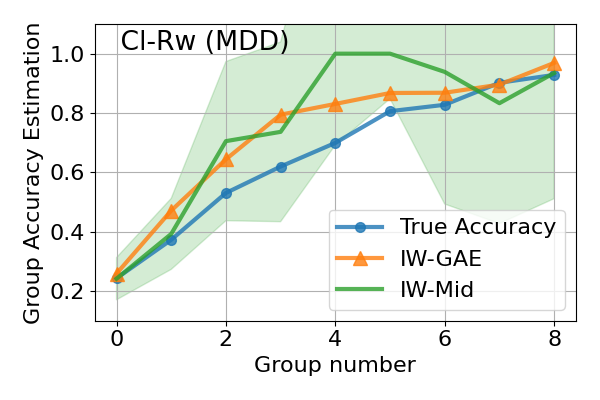}}
    \end{subfigure}
    \begin{subfigure}
     {\includegraphics[width=0.23\textwidth]{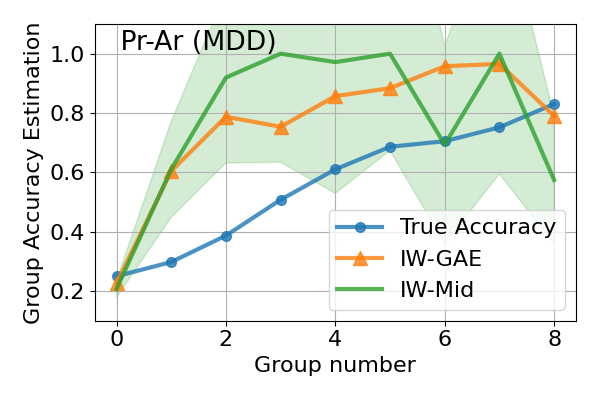}}
    \end{subfigure}
    \begin{subfigure}
     {\includegraphics[width=0.23\textwidth]{experiments/MDD_Pr-Cl.png}}
    \end{subfigure} \\
    \begin{subfigure}
     {\includegraphics[width=0.23\textwidth]{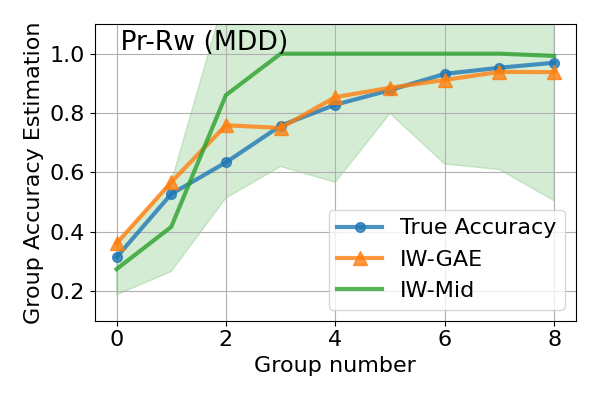}}
    \end{subfigure}
    \begin{subfigure}
     {\includegraphics[width=0.23\textwidth]{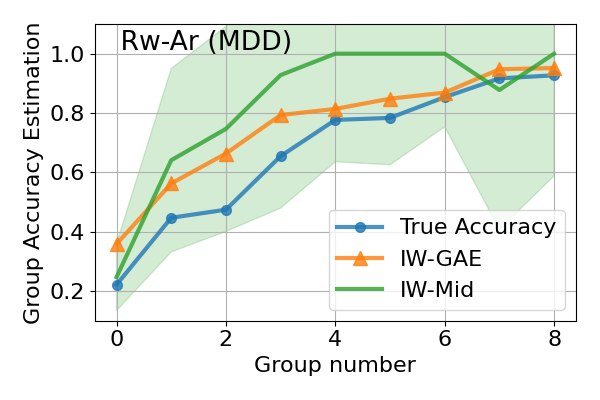}}
    \end{subfigure}
    \begin{subfigure}
     {\includegraphics[width=0.23\textwidth]{experiments/MDD_Rw-Cl.png}}
    \end{subfigure}
    \begin{subfigure}
     {\includegraphics[width=0.23\textwidth]{experiments/MDD_Rw-Pr.png}}
    \end{subfigure}}
    % \hfill
    \caption{True group accuracy and estimated group accuracy of IW-GAE and IW-Mid under MDD. The shaded areas represent possible group accuracy estimation with binned IWs in the CI. The title of a figure represents ``Source-Target.'' For IW-Mid and IW-GAE, we clip the accuracy estimations when they exceed 1, which can occur when the upper bound of CI is large. Also, the number of groups in the figure is different for some domain pairs because there can be a group that contains no target samples (we set $M=10$ for all cases).}
    \label{fig:rc_curve}
    \vskip -0.1in
\end{figure}

To qualitatively analyze IW-GAE, we also visualize reliability curves that compare the estimated group accuracy with the average accuracy in Figure \ref{fig:rc_curve}. We first note that IW-GAE tends to accurately estimate the true group accuracy for most groups under different cases compared to IW-Mid. The accurate group accuracy estimation behavior of IW-GAE explains the results that the IW-GAE improves IW-Mid for most cases in the model calibration and selection tasks (cf. Tables \ref{table:full_office_home_ece}-\ref{table:add_mcd_ece}). For most cases, true accuracy is in between the lower and upper IW estimators, albeit the interval length tends to increase for high-confidence groups. This means that the CI of the IW based on the Clopper-Pearson method successfully captures the IW in the CI. We also note that the true accuracy is close to the lower IW estimator in the lower confidence group and the middle IW estimator in the high confidence group. An observation that the true accuracy's relative positions in CIs varies from one group to another group motivates why an adaptive selection of binned IWs as ours is needed.

\section{Analysis of $\epsilon_{opt}(w^\dagger(n))$ and $IdentBias(w^\dagger(n); \gG_n) $ of IW and Their Relation to Source and Target Group Accuracy Estimation Errors} \label{appx:terms_analy}
In this section, we aim to answer the following question about the central idea of this work: \textit{``Does solving the optimization problem in \eqref{eq:opt_prob}-\eqref{eq:const2} result in an accurate target group accuracy estimator?''} Specifically, we analyze the relationship between the optimization error $\epsilon_{opt}(w^\dagger(n))$, the bias of the identical accuracy assumption $IdentBias(w^\dagger(n); \gG_n) $, the source group accuracy estimation error $|\alpha_S(\gG_n; w^*) - \alpha_S(\gG_n; w^\dagger(n))|$, and the target group accuracy estimation error $|\alpha_T(\gG_n; w^*) - \alpha_T(\gG_n; w^\dagger(n))|$ from the perspective of \eqref{eq:target_acc_estim_bound} and \eqref{eq:tightning_obj}\footnote{Technically speaking, the computed values in this experiment are the empirical expectation which can contain a statistical error. However, since we have no access to the data generating distribution, we perform the analysis as if these values are the population expectations.}. To this end, we gather $w^\dagger(n)$ obtained by solving the optimization problem under all temperature parameters in the search space $t \in \gT$ with MDD on the OfficeHome dataset (720 IWs from 6 values of the temperature parameter, 12 domain pairs, and 10 groups). Then, by using the test dataset in the source and the target domains, we obtain the following observations.

In \eqref{eq:target_acc_estim_bound}, we show that $|\alpha_T(\gG_n; w^*) - \alpha_T(\gG_n; w^\dagger(n))|$ is upper bounded by $|\alpha_S(\gG_n; w^*) - \alpha_S(\gG_n; w^\dagger(n))|$. However, the inequality could be loose since the inequality is obtained by taking the maximum over the IW values. Considering that the optimization problem is formulated for finding $w^\dagger(n)$ that achieves small $|\alpha_S(\gG_n; w^*) - \alpha_S(\gG_n; w^\dagger(n))|$ (cf. Proposition \ref{proposition:closeness_opt}), the loose connection between the source and target group accuracy estimation errors can potentially enlighten a fundamental difficulty to our approach. However, as we can see from Figure \ref{fig:terms_analy}, it turns out that $|\alpha_S(\gG_n; w^*) - \alpha_S(\gG_n; w^\dagger(n))|$ is strongly correlated with $|\alpha_T(\gG_n; w^*) - \alpha_T(\gG_n; w^\dagger(n))|$. This result validates our approach of reducing the source accuracy estimation error of the IW-based estimator for obtaining an accurate group accuracy estimator in the target domain.

\begin{figure}
\centering
    \hfill
    \begin{subfigure}
     {\includegraphics[width=0.4\textwidth]{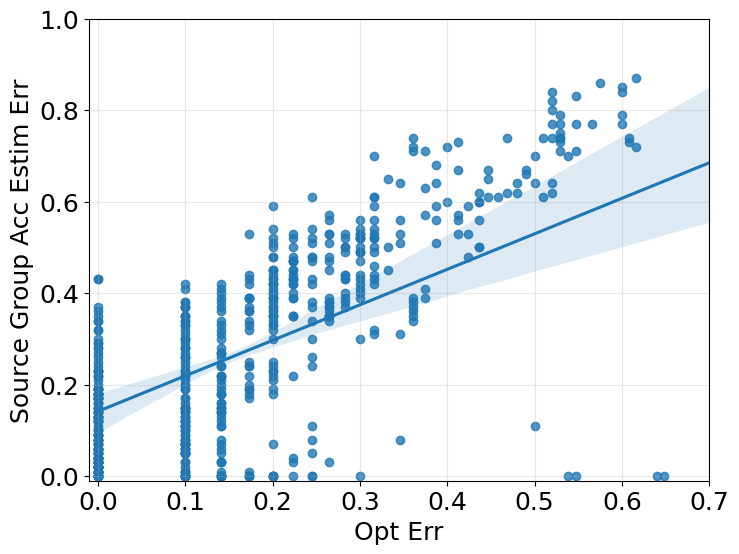}}
    \end{subfigure}
    \hfill
    \begin{subfigure}
     {\includegraphics[width=0.4\textwidth]{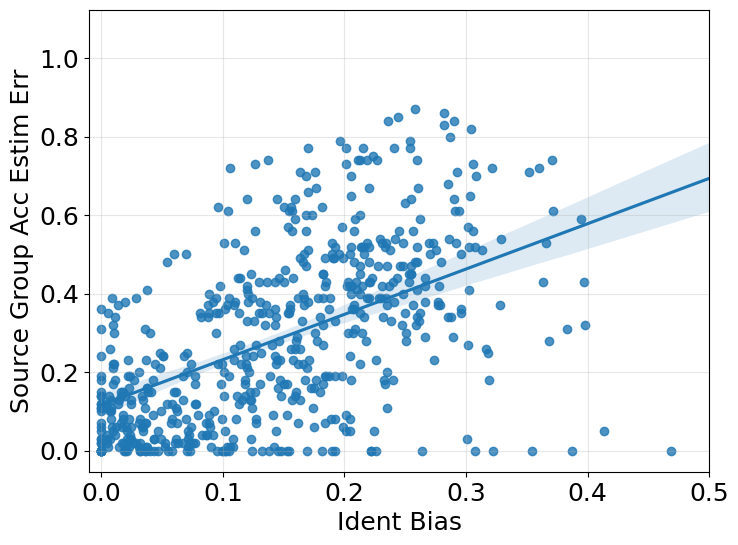}}
    \end{subfigure}
    \hfill
    \caption{
    The relationship between between $\epsilon_{opt}(w^\dagger(n))$ and the source group accuracy estimation error (left) and the relationship between $IdentBias(w^\dagger(n); \gG_n)$ and the source group accuracy estimation error (right). 
    }
    \label{fig:opt_source}
\end{figure}

\begin{figure}
\centering
    \hfill
    \begin{minipage}{.45\textwidth}
      \centering
         {\includegraphics[width=0.9\textwidth]{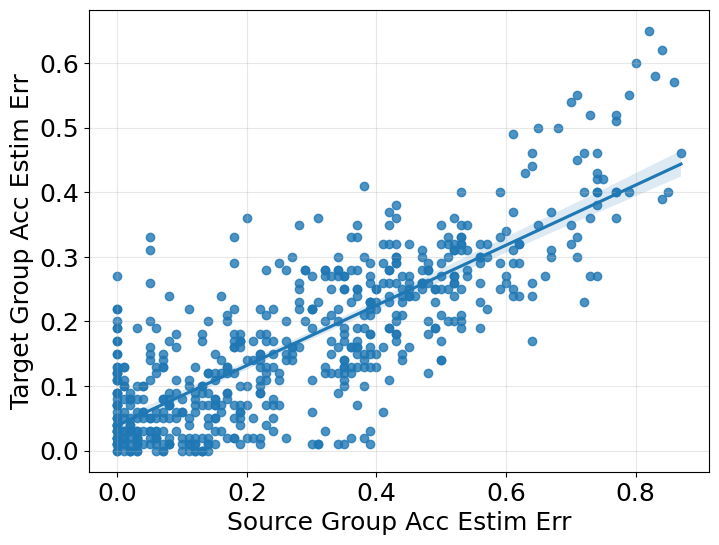}}
      \captionof{figure}{The relationship between source and target group accuracy estimation errors.  }
        \label{fig:terms_analy}
    \end{minipage}%
    \hfill
    \begin{minipage}{.45\textwidth}
      \centering
         {\includegraphics[width=0.9\textwidth]{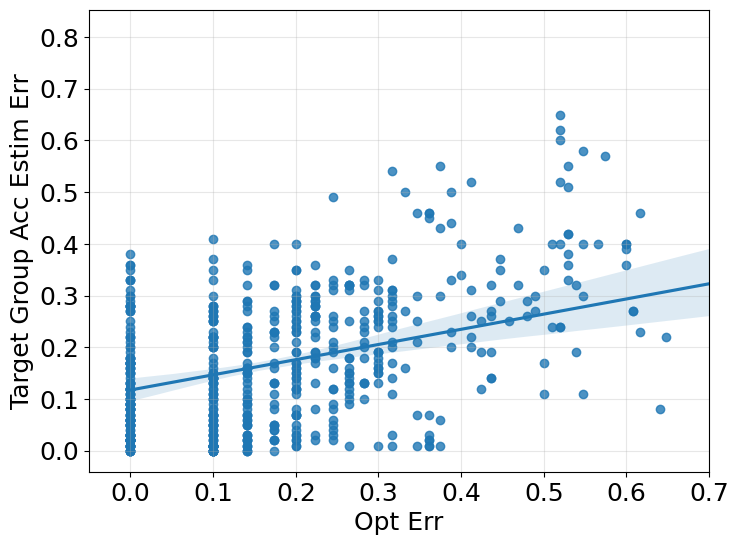}}
      \captionof{figure}{The relationship between $\epsilon_{opt}(w^\dagger(n))$ and the target group accuracy estimation error. }
        \label{fig:opt_target}
    \end{minipage}%
    \hfill
\end{figure}

In \eqref{eq:tightning_obj}, we show that $|\alpha_S(\gG_n; w^*) - \alpha_S(\gG_n; w^\dagger(n))| \leq \epsilon_{opt}(w^\dagger(n)) + \epsilon_{stat} + IdentBias(w^\dagger(n); \gG_n)$, which motivates us to solve the optimization problem for reducing $\epsilon_{opt}(w^\dagger(n))$ (cf. Section \ref{sec:new_method}) and to construct groups based on the maximum value of softmax for reducing $IdentBias(w^\dagger(n); \gG_n)$ (cf. Section \ref{sec:gr_assign}). Again, if these terms are loosely connected to $|\alpha_S(\gG_n; w^*) - \alpha_S(\gG_n; w^\dagger(n))|$, a fundamental difficulty arises for our approach. In this regard, we analyze the relationship between $\epsilon_{opt}(w^\dagger(n))$, $IdentBias(w^\dagger(n); \gG_n)$, and $|\alpha_S(\gG_n; w^*) - \alpha_S(\gG_n; w^\dagger(n))|$. From Figure \ref{fig:opt_source}, we can see that both $\epsilon_{opt}(w^\dagger(n))$ and $IdentBias(w^\dagger(n); \gG_n)$ are strongly correlated to the source group accuracy estimation error. Combined with the observation in Figure \ref{fig:terms_analy}, this observation explains the impressive performance gains by IW-GAE developed for reducing $\epsilon_{opt}(w^\dagger(n))$ and $IdentBias(w^\dagger(n); \gG_n)$.

Next, we analyze the efficacy of solving the optimization problem for obtaining an accurate target group accuracy estimator. To this end, we analyze the relationship between $\epsilon_{opt}(w^\dagger(n))$ and $|\alpha_T(\gG_n; w^*) - \alpha_T(\gG_n; w^\dagger(n))|$. From Figure \ref{fig:opt_target}, $\epsilon_{opt}(w^\dagger(n))$ is correlated with $|\alpha_T(\gG_n; w^*) - \alpha_T(\gG_n; w^\dagger(n))|$, which explains the performance gains in the model calibration and selection tasks by IW-GAE. However, the correlation is weaker than the cases analyzed in Figure \ref{fig:terms_analy} and Figure \ref{fig:opt_source}. We conjecture that this is because $\epsilon_{opt}(w^\dagger(n))$ is connected to $|\alpha_T(\gG_n; w^*) - \alpha_T(\gG_n; w^\dagger(n))|$ through two inequalities \eqref{eq:target_acc_estim_bound} and \eqref{eq:tightning_obj}, and this results in a somewhat loose connection between $\epsilon_{opt}(w^\dagger(n))$ and $|\alpha_T(\gG_n; w^*) - \alpha_T(\gG_n; w^\dagger(n))|$.

In Figure \ref{fig:opt_target}, we also note that the optimization problem is subject to a non-identifiability issue that the solutions with the same optimization error can have significantly different target group accuracy estimation errors (e.g., points achieving the zero optimization error in Figure \ref{fig:opt_target}). We remark that the non-identifiability issue motivates an important future direction of research that develops a more sophisticated objective function and a regularization function that can distinguish estimators with different target group accuracy estimation errors.

\end{document}